%% file: iclr2027_conference.tex
\documentclass{article} 
\usepackage{iclr2027_conference,times}

\input{math_commands.tex}

\usepackage{url}
\usepackage{booktabs}
\usepackage{graphicx}
\usepackage{multirow}
\usepackage{siunitx}
\usepackage{adjustbox}
\usepackage{amssymb}
\usepackage{arydshln}
\usepackage{subcaption}
\usepackage{hyperref}
\usepackage{algorithm}
\usepackage{algorithmic}
\usepackage{multicol}
\usepackage{wrapfig}
\usepackage{xcolor}
\usepackage[breakable]{tcolorbox}
\usepackage{listings}
\usepackage{colortbl}
\definecolor{harnessbg}{HTML}{CCE6FF}
\usepackage{xspace}
\DeclareUnicodeCharacter{266B}{}

\definecolor{toolcardbox}{RGB}{240, 248, 255}
\definecolor{toolcardborder}{RGB}{52, 52, 173}

\newtcolorbox{custombox}[2][]{
    colback=toolcardbox,
    colframe=toolcardborder,          
    coltitle=white,                
    arc=1pt,                       
    boxrule=1pt,
    fonttitle= ,
    title=#1,                      
    left=5pt,
    right=5pt,
    top=5pt,
    bottom=5pt,
    before skip=1em,
    after skip=1em,
    fontupper=\small,               
    breakable,     
    width=1.\linewidth, 
    #1                        
}

\NewDocumentCommand{\shanyong}
{ mO{} }{\textcolor{blue}{\textsuperscript{\textit{Shanyong}}\textsf{\textbf{\small[#1]}}}}

\NewDocumentCommand{\yanyu}
{ mO{} }{\textcolor{yellow}{\textsuperscript{\textit{Yanyu}}\textsf{\textbf{\small[#1]}}}}

\NewDocumentCommand{\zhenwen}
{ mO{} }{\textcolor{purple}{\textsuperscript{\textit{Zhenwen}}\textsf{\textbf{\small[#1]}}}}

\newcommand{\ourmethod}{Harness-Search\xspace}

\title{Harness-Search: Guiding Long-Horizon Search through Multi-Agent Coordination}

\author{Shanyong Wang$^{1}$, Zhenwen Ji$^{1,2}$, Lei Jin$^1$, Yining Zhao$^3$, Yicheng Qian$^1$ \\
\textbf{Chengqiang Lu$^1$,Yi Wu$^1$, Yao Hu$^1$, Lizhen Cui$^2$, Yanyu Xu$^2$} \\
$^1$Xiaohongshu Inc. \\
$^2$the Joint SDU-NTU Centre for Artificial Intelligence Research (C-FAIR), Shandong University \\
$^3$University of Illinois at Urbana-Champaign \\
\texttt{wangshanyong@xiaohongshu.com, xu\_yanyu@sdu.edu.cn} \\
}

\iclrfinalcopy 
\begin{document}

\maketitle

\begin{abstract}
Long-horizon search requires agents to gather evidence across multiple steps and synthesize it into well-supported answers. 
The recent agent harnesses provide a natural and promising framework to support such long-running search processes.
As interaction histories grow, one single agent in harnesses might get stuck and cause the policy to lose track of unresolved questions, overlook useful evidence, or terminate before sufficient support has been collected. One of promising way is to decouple three distinct responsibilities of proposing retrieval actions, updating persistent state, and deciding when to stop rather than concentrating them within a single policy.
Targeted at it, we introduce \ourmethod, a multi-agent search harness to reduce the local errors propagating across subsequent exploration, evidence curation, and termination decisions.
In particular, \ourmethod assigns these responsibilities to three permission-bounded authorities: a Retrieval Policy that proposes search operations, a Memory Operator that validates and commits persistent-state updates, and a Summary Auditor that accepts or rejects termination based on the sufficiency of the curated evidence.
Together, these roles form a Propose–Commit–Audit loop in which actions are proposed, persistent evidence is selectively committed, and stopping decisions are subjected to an explicit sufficiency check.
Across seven long-horizon search benchmarks, \ourmethod improves both retrieval and answer generation under the same policy backbone, increasing Recall by 4.60--27.92 points and Final-Answer Recall by 12.34--30.13 points over the strongest harness-based baseline on each evidence-retrieval benchmark. Moreover, trajectory-level analyses show that \ourmethod continues to accumulate useful evidence and expand evidence coverage with less redundant retrieval as the search history grows. 
Our code is available at \href{https://github.com/SwimmingWang/Harness-Search}{https://github.com/SwimmingWang/Harness-Search}.
\end{abstract}

\begin{figure}[t!]
    \centering
    \includegraphics[width=0.96\linewidth]{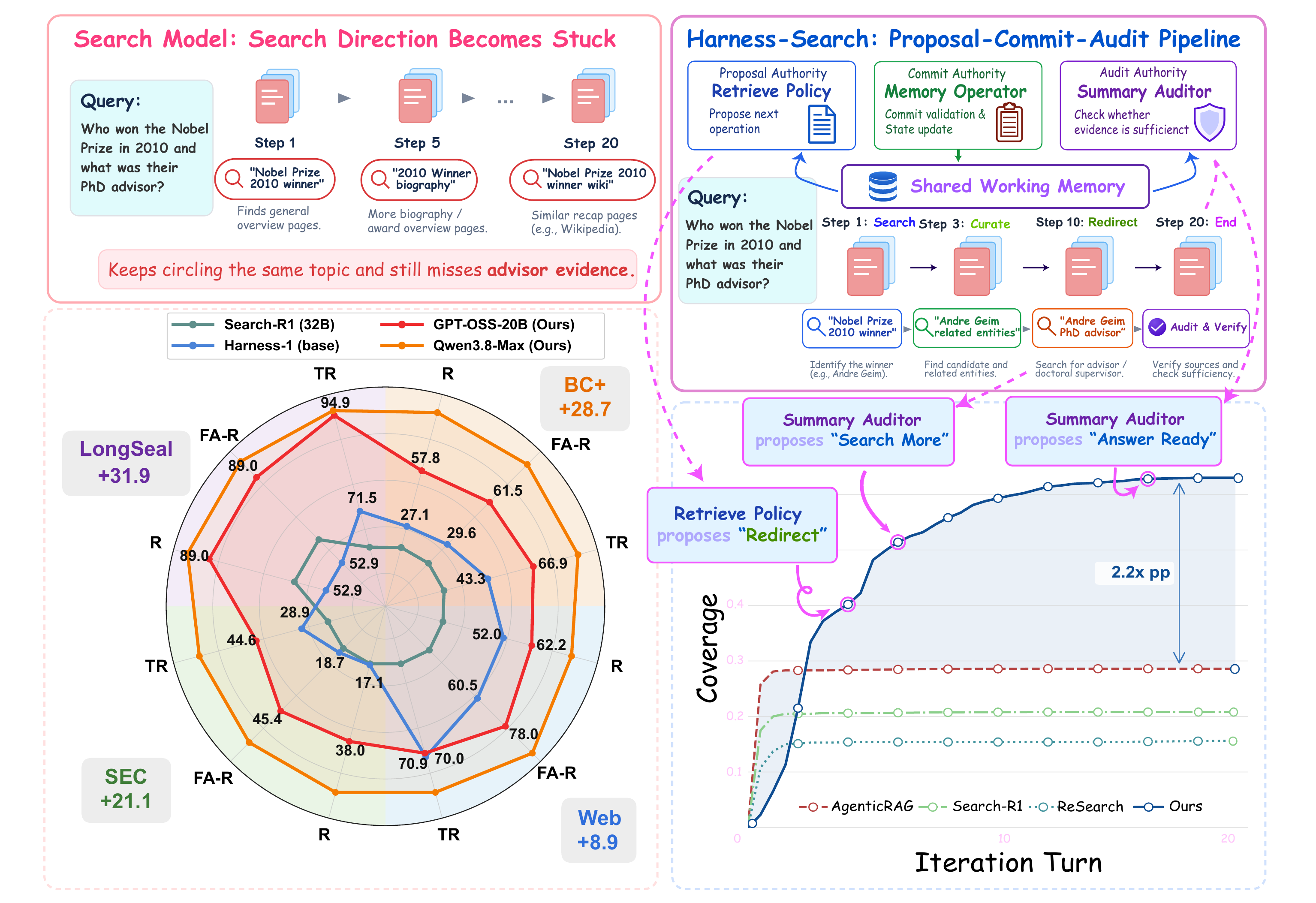}
    \caption{\textbf{Harness-Search} turns open-ended search into a guided, evidence-driven process. Naive iterative search often gets stuck exploring the same direction, leading to redundant retrieval and missing evidence. Harness-Search explicitly alternates between planning, searching, curating, and auditing, and redirects the search when the auditor identifies an evidence gap. As a result, it continually expands search coverage instead of saturating early (\textbf{Bottom Right}), yielding substantial improvements across diverse search-intensive benchmarks (\textbf{Bottom Left}).}
    \label{fig:intro_figure}
\end{figure}

\section{Introduction}

\input{section/1_introduction}

\section{Related Work}
\input{section/2_related_work}
\begin{figure}[t]
    \centering
    \includegraphics[width=0.96\linewidth]{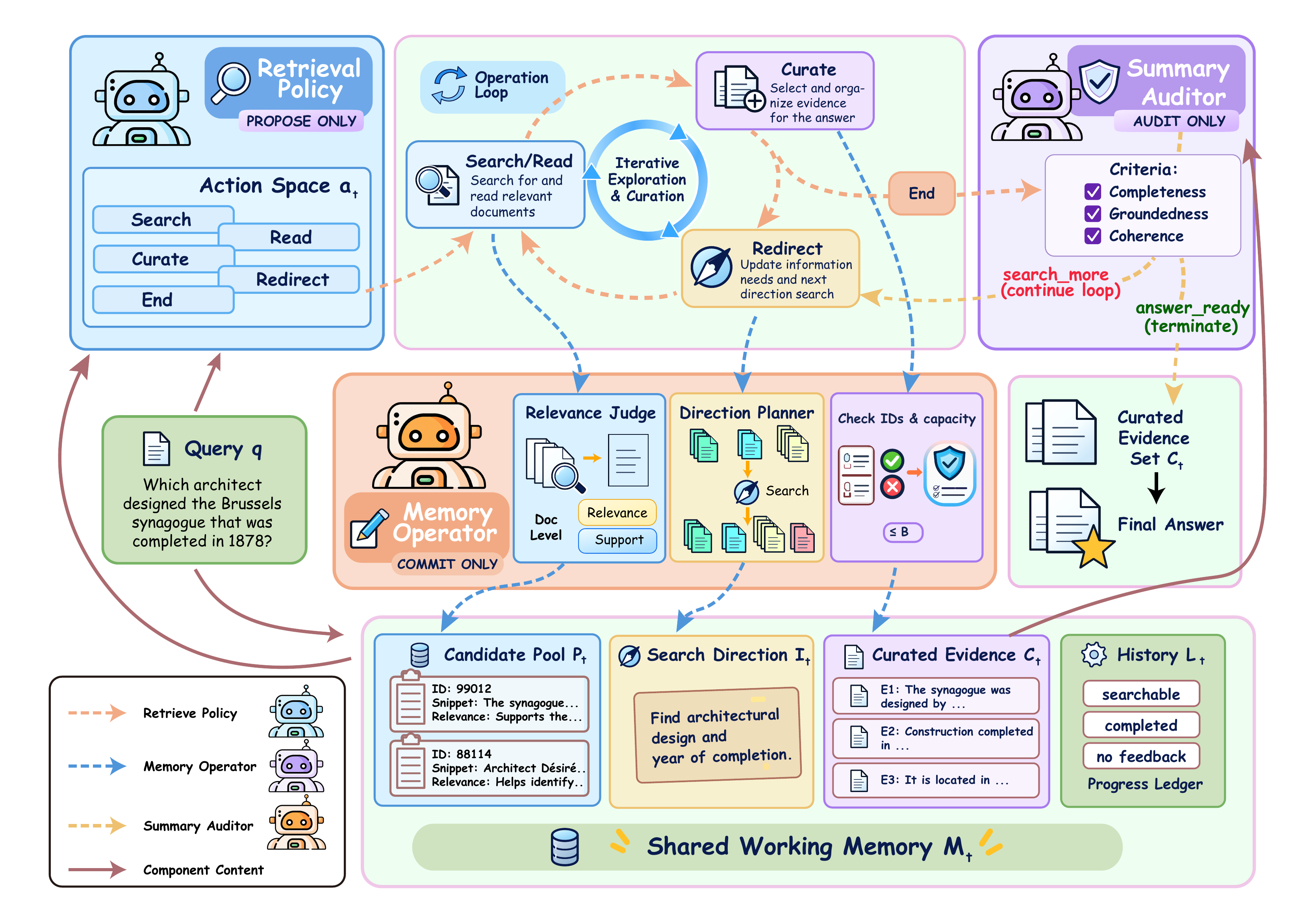}
    \caption{Overview of \textbf{\ourmethod}. \ourmethod addresses long-horizon search with three permission-bounded agents: a Retrieval Policy proposes actions, a Memory Operator validates and commits persistent state, and a Summary Auditor audits evidence sufficiency before termination.}
    \label{fig:main_figure}
\end{figure}

\section{Method}
\label{sec:method}
\input{section/3_method}

\section{Main Experiments}
\input{section/4_experiment}

\section{Analysis}
\input{section/5_analysis}

\section{Conclusion}
We presented \ourmethod, a multi-agent retrieval harness that coordinates a Retrieval Policy, Memory Operator, and Summary Auditor through shared working memory, separating action proposal, persistent evidence management, and termination auditing within a long-horizon search loop. By organizing search as a Propose--Commit--Audit process, \ourmethod enables iterative exploration while preserving useful evidence and explicitly checking whether the accumulated evidence is sufficient before termination. Comprehensive experiments across evidence-retrieval and multi-hop QA benchmarks show that \ourmethod consistently improves evidence discovery and retention over existing retrieval and harness-based systems, with gains transferring across policy models of different scales and kinds. Further analyses demonstrate the complementary roles of memory management, relevance assessment, and termination auditing, and show that the harness can also be used for supervised fine-tuning.

\section*{Ethics statement}

We conduct fundamental research on Retrieval-Augmented Generation and tool use in large language models. Our method is developed and evaluated exclusively on publicly available benchmarks, without involving human subjects or using sensitive personal data or private information. The proposed training pipeline is designed as a general-purpose optimization technique. Within the scope of this study, we do not foresee specific ethical risks related to fairness, bias, discrimination, privacy, or security. This research is conducted in accordance with established standards of research integrity.

\section*{AI Use Statement}
\label{sec:ai}
AI systems, including LLMs, did not contribute to the development of the research ideas or the writing of this paper to a degree that would warrant authorship or contributor status. In this work, LLMs are treated strictly as objects of study and are the primary focus of our experimental evaluations.

\section*{Limitation}
Despite its effectiveness, \ourmethod has several limitations. Multi-agent coordination and repeated auditing can introduce additional inference overhead, creating a trade-off between evidence quality and efficiency. Performance also depends on the models used by the Memory Operator and Summary Auditor, and rejecting a termination proposal does not guarantee that subsequent search will recover missing evidence. Finally, our evaluation focuses on evidence-retrieval and multi-hop QA benchmarks; generalization to dynamic open-web environments and broader information-seeking tasks remains to be established.

\bibliography{iclr2027_conference}
\bibliographystyle{iclr2027_conference}

\appendix
\newpage

\input{section/appendix}

\end{document}

%% file: math_commands.tex
\usepackage{amsmath,amsfonts,bm}

\def\eqref#1{equation~\ref{#1}}

\def\1{\bm{1}}

\DeclareMathAlphabet{\mathsfit}{\encodingdefault}{\sfdefault}{m}{sl}
\SetMathAlphabet{\mathsfit}{bold}{\encodingdefault}{\sfdefault}{bx}{n}



%% file: section/1_introduction.tex
Long-horizon search requires agents to gather evidence across multiple steps and synthesize it into well-supported answers. 
Standard search agents typically handle this process with a single policy that interleaves reasoning and retrieval until it decides to produce a final answer~\citep{yao2022react,trivedi2023interleaving,asai2024self,chen2025decisionflow}. 
Crucially, the intermediate findings may reveal new constraints, resolve ambiguous entities, or expose missing links. In other wards, what should be searched for next often depends on what has already been found~\citep{trivedi2022musique,yang2018hotpotqa,ho2020constructing}.
Successful long-horizon search therefore depends not only on retrieving relevant documents, but also on maintaining a coherent search state and retaining useful evidence throughout the trajectory.




Very recent agent harnesses provide a natural and promising framework to support such long-running search processes~\citep{jiang2026harness, wu2026memharness}. 
In particular, a harness acts as an external execution and control layer around an agent, organizing how the model interacts with tools, maintains persistent state, and coordinates multi-step workflows. 
Compared with relying on a single model call or a fixed retrieve-then-generate pipeline, a harness can therefore sustain search over much longer trajectories while exposing intermediate decisions to explicit control. 
The recent harness based long-horizon search methods allows an agent to repeatedly invoke retrieval and reading tools~\citep{sen2026grep}, preserve information discovered in earlier stages through memory~\citep{huang2026memoharness}, track its current search state~\citep{ma2026longhorizon}, and adapt subsequent actions based on accumulated evidence~\citep{zhang2026scaling}.
The above methods make extended search more manageable by externalizing information that would otherwise be buried in the interaction history. 




One single policy may move on while an information need remains unresolved, fail to retain a useful document it has already found, or stop before the retained evidence supports an answer. We refer to the concentration of these decisions within a single policy as \emph{authority coupling}: the policy proposes retrieval actions, determines what persists in the answer-facing evidence set, and judges when the search is complete. 
First, accumulated findings may not be organized into an explicit account of search coverage and unresolved information needs. Without a clear distinction between what has been established and what remains missing, the agent may repeatedly search familiar directions rather than pursue unmet evidence needs. For example, Figure~\ref{fig:intro_figure} (top left) illustrates an agent asked to identify a 2010 Nobel laureate and their PhD advisor. It repeatedly retrieves winner lists and general biographies instead of shifting its focus to the unresolved advisor relationship. Second, useful evidence may become buried in the expanding interaction history~\citep{liu2024lost, du2025context}. Even when a relevant document has already been retrieved, the agent may fail to recover and use it in subsequent steps or omit it from the final answer context. Third, even when the agent believes that sufficient evidence has been collected, the retained pieces may fail to form a coherent evidence chain when synthesized together. Required facts may still be missing, logical connections may remain unsupported, or individually relevant documents may not jointly support the final conclusion.
This motivates separating proposal, state commitment, and termination audit into distinct control authorities.

Thus, we introduce \textbf{\ourmethod}, a multi-agent search harness that explicitly separates the control of evidence discovery, persistent evidence management, and termination verification. \ourmethod organizes search as a \emph{Propose--Commit--Audit} loop with three permission-bounded roles. Inside the system, a \textbf{Retrieval Policy} proposes search, reading, redirection, curation, and termination operations based on an explicit representation of the current retrieval direction and prior search progress, adaptively guiding the retrieval process.
A \textbf{Memory Operator} exclusively manages persistent working memory, maintaining the discovered evidence pool, curated answer-facing evidence, active retrieval intent, and progress ledger while exposing only a bounded rendering of this state to the Retrieval Policy. Finally, an \textbf{Summary Auditor} treats termination as a proposal rather than an unconditional action: it attempts to synthesize the curated evidence and verifies its completeness, groundedness, and coherence. When the evidence is insufficient, the identified information or logical gap is written back into memory as the target of further retrieval. Together, these components establish an explicit evidence lifecycle in which search coverage is expanded, useful evidence is selectively retained, and stopping is conditioned on whether the accumulated evidence can actually support a coherent answer.

We evaluate \ourmethod on seven long-horizon search benchmarks, including four evidence-retrieval benchmarks and three multi-hop answer-generation benchmarks. With GPT-OSS-20B as the policy model, \ourmethod consistently improves the quality of the final evidence set over existing harness-based systems, increasing Recall by 4.60--27.92\% and Final-Answer Recall by 12.34--30.13\% relative to the strongest harness baseline on each retrieval benchmark. These improvements further translate to answer generation, where \ourmethod outperforms compared harness-based methods under the same policy backbone. Beyond the main benchmark results, we further investigate how \ourmethod behaves throughout the long-horizon search process. We find that, as the histories increases, \ourmethod continues to accumulate useful evidence and improve evidence coverage while reducing redundant retrieval (Figure~\ref{fig:intro_figure}). We also explore whether the performance can be further improved through model training. By applying supervised fine-tuning (SFT) to the policy model, we observe additional gains over the original backbone. 

To summarize, our paper’s contributions are threefold:
\begin{itemize}
    \item We introduce \ourmethod, a multi-agent search harness that decouples search control, persistent evidence curation, and termination auditing;
    \item We show that the harness improves evidence discovery and retention across heterogeneous long-horizon retrieval benchmarks and transfers across policy models;
    \item We analyze the search dynamics of \ourmethod over long-horizon trajectories, showing that it continues to accumulate useful evidence and expand evidence coverage as interaction turns increase, while reducing redundant searches and avoiding stagnation around previously explored retrieval directions.
\end{itemize}

%% file: section/2_related_work.tex
\textbf{Long-Horizon Search.} 
Long-horizon search tasks~\citep{yang2018hotpotqa, trivedi2022musique,pham2026sealqa} require repeated interaction between a model and its environment. Prior work approaches this problem from several directions. Reasoning-and-retrieval methods interleave reasoning with information acquisition. ReAct~\citep{yao2022react}, for example, introduced a reason-act-observe loop in which agents reason about their current state, take actions, and update their reasoning based on new observations. Later retrieval-focused methods, including IRCoT~\citep{trivedi2023interleaving}, Iter-RetGen~\citep{shao2023enhancing}, FLARE~\citep{jiang2023active}, and Self-RAG~\citep{asai2024self}, were inspired by this. More recent agentic retrieval methods extend this approach to longer search trajectories, allowing models to adjust queries and tool calls in response to newly acquired evidence~\citep{li2025search,chen2025research,zheng2025deepresearcher,wang2026meet,yu2025sotopia,yu2026ctm,ji2026momentum}. Moreover memory-centric methods provide agents with external or structured memory for storing, organizing, and retrieving information accumulated over long trajectories~\citep{packer2023memgpt, chhikara2025mem0, xu2026mem,ye2026auto}. As search processes grow longer, effective agents increasingly depend not only on reasoning but also on memory management and tool use to preserve state and interact with external environments.

\textbf{Harness engineering.} 
Harness-based approaches provide a largely training-free way to shape agent behavior, making them easy to deploy, modify, and transfer across tasks. A growing range of systems~\citep{yang2024swe,wu2023autogen,zhang2025aflow,packer2023memgpt,fourney2024magentic,pan2026natural} use external execution structures to impose specialized roles, route tool use, maintain task state, or coordinate multi-step workflows. Such explicit control improves modularity and makes agent behavior easier to revise without retraining, a property also exploited in long-running agent setups~\citep{lee2026metaharnessendtoendoptimizationmodel} and recent work on adaptive or automatically evolving harnesses~\citep{huang2026memoharness,park2026autosaddler,wu2026harnessdev}. However, richer execution structures can also induce longer interaction traces and greater state-management overhead: as intermediate actions, observations, and histories accumulate, maintaining the relevant context becomes increasingly difficult, and the model may lose track of earlier information or the current objective~\citep{hu2025hiagent,liu2024lost,ye2025agentfold}.

%% file: section/3_method.tex

\subsection{Problem Formulation}
\label{sec:problem_formulation}

Given a query $q$ and a document corpus $\mathcal{D}$, the goal of long-horizon retrieval is to construct a bounded evidence set $\mathcal{C}_T \subseteq \mathcal{D}$ that is sufficient to support the query.
Let $\mathcal{R}_t$ denote the documents returned by retrieval at turn $t$. We use $\mathcal{P}_t$ to denote all documents encountered up to turn $t$, and $\mathcal{C}_t$ to denote the bounded set of evidence retained for final answer generation, where $\mathcal{C}_t \subseteq \mathcal{P}_t$ and $\vert \mathcal{C}_t \vert \leq B$, where $B$ is the maximum evidence capacity. For tasks that require the generation of answers, the system additionally returns an evidence-based synthesis constructed from $\mathcal{C}_T$.

\ourmethod{} aims to separate three decisions that are often handled by a single search agent: what to do next, what information should persist, and whether the available evidence is sufficient to stop. It contains a \textbf{Retrieval Policy}, a \textbf{Memory Operator}, and a \textbf{Summary Auditor}, as shown in Figure~\ref{fig:main_figure}. The Retrieval Policy proposes retrieval operations, the Memory Operator is responsible for persistent state updates, and the Summary Auditor decides whether a proposed termination is supported by the curated evidence.

\subsection{Separation of Control Authorities}
\label{sec:harness_search}

It separates three control authorities that are commonly coupled within a single retrieval trajectory. The \textbf{Retrieval Policy} determines what operation should be attempted next. The \textbf{Memory Operator} is the only component allowed to commit changes to working memory. The \textbf{Summary Auditor} independently determines whether a termination proposal is supported by the curated evidence.

\textbf{Shared working memory.} 
The three components coordinate through a shared working memory, $
\mathcal{M}_t =
\left(
q,\,
\mathcal{P}_t,\,
\mathcal{C}_t,\,
I_t,\,
\mathcal{L}_t
\right),
$ where $\mathcal{P}_t$ contains all documents encountered so far, $\mathcal{C}_t$ is the curated evidence retained for answer generation, $I_t$ represents the current search direction, and $\mathcal{L}_t$ is a history recording explored, unsuccessful, and unresolved directions.

\textbf{Execution budgets.}
The interaction among three authorities is constrained by two execution budgets. $T_{\max}$ denotes the maximum number of policy-operation turns in a trajectory and $S_{\max}$ is the maximum number of \texttt{search} operations that may be executed between two curation steps. 

\subsection{Proposal Authority: Retrieval Policy}

At each turn, the Retrieval Policy selects one operation from action space $a_t$: \{\texttt{search}, \texttt{read}, \texttt{redirect}, \texttt{curate}, or \texttt{end}\}. \texttt{search} retrieves candidate documents and \texttt{read} obtains additional content from a selected document. \texttt{redirect} indicates that the current search direction should be revised, while \texttt{curate} proposes changes to the answer-facing evidence set. Finally, \texttt{end} proposes that the current evidence is sufficient to terminate retrieval.

Unlike a conventional fixed action space, \ourmethod{} dynamically restricts the admissible actions according to the current search state. Let \(s_t\) denote the number of \texttt{search} operations executed since the most recent \texttt{curate}. The action space is restricted as
\begin{equation}
\mathcal{A}_t
=
\begin{cases}
{\texttt{curate}}, & \text{if } s_t \geq S_{\max}, \\[2mm]
{\texttt{redirect}, \texttt{end}}, & \text{if } a_{t-1}=\texttt{curate}.
\end{cases}
\label{eq:admissible_actions}
\end{equation}
When neither condition is triggered, all actions in \(\mathcal{A}\) remain admissible. After \(S_{\max}\) consecutive search operations, the policy is required to \texttt{curate} the retrieved evidence before taking further retrieval actions. Based on the curated evidence, it must then either \texttt{redirect} to a new retrieval direction or \texttt{end} the search if the current evidence is already sufficient. This mechanism periodically reassesses whether further exploration along the current direction is necessary, preventing the policy from repeatedly searching along an unproductive direction and mitigating cases where sufficient evidence has already been retrieved but is not yet recognized by the retrieval policy.



\subsection{Commit Authority: Memory Operator}


\textbf{Committing retrieval observations.} For \texttt{search} and \texttt{read}, newly encountered documents are deduplicated and incorporated into the candidate pool. The Memory Operator additionally invokes a document-level \textbf{Relevance Judge} to construct query-relative memory annotations. 

\textbf{Committing direction and progress updates.} 
For \texttt{redirect}, the Memory Operator invokes its internal \textbf{Direction Planner} to construct a revised retrieval direction. After validating the selected document identifiers and processing the returned structure, the operator commits the update as $I_{t+1}$ and records the previous searching direction and policy rationale in history.
 

\textbf{Committing curated evidence.}
For \texttt{curate}, the operator validates the proposed additions and removals before updating $\mathcal{C}_t$, rejecting invalid document identifiers, duplicate operations, or updates that exceed the evidence budget $B$.

The distinction between proposal and commitment prevents policy-generated content from being treated as trusted persistent state by default. At the same time, the Memory Operator remains a constrained state manager: it validates and records the consequences of policy decisions but does not independently choose the next retrieval operation or approve termination.

\subsection{Audit Authority: Evidence-Gated Termination}

An \texttt{end} operation is treated as a termination proposal rather than an immediate stop. Given the current curated evidence, the Summary Auditor first attempts to synthesize an answer $y_t$ to the query. It then uses this synthesis process to assess whether the available evidence is sufficient, considering three aspects: \emph{completeness}, whether the information needed to answer the query is present; \emph{groundedness}, whether the generated answer is adequately supported by the retained evidence; and \emph{coherence}, whether the evidence can be connected into a consistent answer.

Based on this assessment, the auditor returns one of two outcomes: \texttt{answer\_ready} or \texttt{search\_more}. If the attempted synthesis can be adequately supported by the current evidence, \texttt{answer\_ready} accepts the termination proposal. Otherwise, \texttt{search\_more} identifies the missing information or unsupported connection exposed during synthesis, and the Retrieval Policy is restricted to \texttt{redirect} in the next turn to revise the retrieval direction accordingly.

Algorithm and prompt are presented in Appendix~\ref{app:algorithm},~\ref{app:prompt}, representatively.



%% file: section/4_experiment.tex
\subsection{Experimental Setup}
\textbf{Benchmarks.}
We evaluate our method on two categories of long-horizon search benchmarks: Evidence-Retrieval and Answer-Generation QA. The Evidence-Retrieval benchmarks include BrowseComp-plus~\citep{chen2025browsecomp}, Web synthetic, SEC filings~\citep{bashir2026context1}, and LongSealQA~\citep{pham2026sealqa}. The Answer-Generation QA benchmarks include HotpotQA~\citep{yang2018hotpotqa}, 2WikiMultiHopQA~\citep{ho2020constructing}, and Musique~\citep{trivedi2022musique} and we utilize the 2018 Wikipedia dump~\citep{karpukhin2020dense} as the foundational knowledge corpus. These two categories differ in their expected outputs. Answer-Generation QA requires the model to generate an answer for each query, while Evidence-Retrieval only requires it to return a set of documents containing evidence relevant to the query.

\begin{table*}[t]
  \caption{
    Main results on Evidence-Retrieval benchmarks. We report Recall (R), Final-Answer Recall (FA-R), and Trajectory Recall (TR) across four benchmarks. \textbf{Bold} marks the best result in each column within each same-backbone comparison including ties. Shaded rows denote \ourmethod.
    $^\dagger$Top-$k$ denotes retrieving $k$ documents per search call.
  }
  \label{tab:recall-results}

  \centering
  \small
  \setlength{\tabcolsep}{3.5pt}
  \renewcommand{\arraystretch}{1.12}

  \sisetup{
    mode = text,
    detect-weight = true
  }

  \begin{adjustbox}{max width=\textwidth}
  \begin{tabular}{
    @{}
    l
    *{12}{S[table-format=2.2]}
    @{}
  }
    \toprule

    \multirow{2}{*}{\textbf{Method}}
      & \multicolumn{3}{c}{\textbf{BC+}}
      & \multicolumn{3}{c}{\textbf{Web}}
      & \multicolumn{3}{c}{\textbf{SEC}}
      & \multicolumn{3}{c}{\textbf{LongSeal}}
      \\

    \cmidrule(lr){2-4}
    \cmidrule(lr){5-7}
    \cmidrule(lr){8-10}
    \cmidrule(l){11-13}

      & {R} & {FA-R} & {TR}
      & {R} & {FA-R} & {TR}
      & {R} & {FA-R} & {TR}
      & {R} & {FA-R} & {TR}
      \\
    \midrule
    \multicolumn{13}{@{}l@{}}{
      \textbf{A. RAG models with native backbones}
    } \\
    \addlinespace[2pt]

    s3 (7B)~\citep{jiang2025s3}
      & 19.29 & 19.29 & 24.48
      & 40.26 & 54.06 & 56.26
      & 22.74 & 22.74 & 27.49
      & 27.16 & 27.16 & 45.80
      \\

    Search-R1 (32B)~\citep{jin2025search}
      & 15.36 & 15.36 & 20.54
      & 30.22 & 30.22 & 47.92
      & 16.81 & 16.81 & 19.82
      & 62.66 & 62.66 & 62.66
      \\

    ReSearch (32B)~\citep{chen2025research}
      & 12.91 & 12.91 & 15.49
      & 28.13 & 28.13 & 38.48
      & 15.89 & 15.89 & 18.71
      & 68.56 & 68.56 & 68.56
      \\

    Tongyi DR (30B)~\citep{tongyidr}
      & 48.97 & 48.97 & 52.10
      & 26.01 & 26.01 & 28.05
      & 22.92 & 22.92 & 25.13
      & 87.63 & 87.63 & 87.63
      \\
      
    \midrule
    \multicolumn{13}{@{}l@{}}{
      \textbf{B. Same-backbone comparison: GPT-OSS-20B}
    } \\
    \addlinespace[2pt]

    RRR (top-10)$^\dagger$
      & 8.32  & 11.98 & 23.09
      & 21.19 & 45.94 & 58.73
      & 8.29  & 12.90 & 24.08
      & 25.19 & 25.19 & 57.87
      \\

    RRR (top-30)$^\dagger$
      & 10.41 & 14.85 & 32.10
      & 22.23 & 47.23 & 69.78
      & 9.25  & 13.86 & 31.17
      & 28.34 & 28.34 & 59.05
      \\

    ACM~\citep{li2026acm}
      & 29.89 & 31.38 & 29.90
      & 40.41 & 52.20 & 40.41
      & 21.91 & 25.35 & 21.91
      & 84.38 & 61.55 & 84.38
      \\

    AgenticRAG~\citep{suresh2026agenticrag}
      & 23.48 & 23.48 & 23.48
      & 36.06 & 36.06 & 36.06
      & 17.83 & 17.83 & 17.83
      & 76.64 & 76.64 & 76.64
      \\

    Harness-1~\citep{jiang2026harness}
      & 27.07 & 29.61 & 43.30
      & 52.00 & 60.47 & \bfseries 70.92
      & 17.13 & 18.74 & 28.95
      & 52.88 & 52.88 & 71.45
      \\

    MemHarness~\citep{wu2026memharness}
      & 25.43 & 25.43 & 29.74
      & 37.24 & 37.24 & 41.45
      & 21.18 & 21.18 & 27.63
      & 64.04 & 64.04 & 64.04
      \\

    \rowcolor{gray!10}
    \textbf{\ourmethod}
      & \bfseries 57.81 & \bfseries 61.51 & \bfseries 66.86
      & \bfseries 62.19 & \bfseries 77.98 & 70.03
      & \bfseries 37.97 & \bfseries 45.44 & \bfseries 44.65
      & \bfseries 88.98 & \bfseries 88.98 & \bfseries 94.88
      \\
    \midrule
    \multicolumn{13}{@{}l@{}}{
      \textbf{C. \ourmethod with additional backbones}
    } \\
    \addlinespace[2pt]
    \rowcolor{gray!10}
    Qwen3.5-27B
      & 69.98 & 72.62 & 79.98
      & 80.33 & 94.01 & 95.76
      & 49.47 & 56.63 & 57.97
      & 88.65 & 88.65 & 95.42
      \\

    \rowcolor{gray!10}
    Qwen3.8-Max
      & 70.12 & 79.89 & 83.23
      & 76.51 & 94.84 & 79.72
      & 51.87 & 59.80 & 64.42
      & 95.67 & 95.67 & 96.06
      \\

    \rowcolor{gray!10}
    Gemini-3.1-pro
      & 79.48 & 84.31 & 86.16
      & 88.93 & 92.09 & 96.57
      & 57.43 & 61.68 & 68.45
      & 95.28 & 95.28 & 98.82
      \\

    \rowcolor{gray!10}
    GPT-5.6-Terra
      & 82.27 & 87.53 & 90.12
      & 77.96 & 79.43 & 98.86
      & 62.82 & 67.97 & 77.62
      & 98.43 & 98.43 & 98.82
      \\

    \bottomrule
  \end{tabular}
  \end{adjustbox}
\end{table*}

\textbf{Baselines.}
We compare against three families of retrieval systems. First, \emph{Rewrite--Retrieve--Rerank} baselines use the policy model to formulate search queries, retrieve candidate documents, and rerank the resulting set. Second, we compare against existing retrieval-augmented and agentic-search methods, including s3, Search-R1, ReSearch, Tongyi DeepResearch, Prompt-R1, KiRAG, StepSearch, GlobalRAG, RECOMP, BAVT, and IRCoT. Third, we compare against harness- and agent-based systems, including ACM, AgenticRAG, Harness-1, MemHarness, and Harness-RL. Additional implementation and reproduction details are provided in Appendix~\ref{app:baselines}.

\textbf{Evaluation metrics.}
For Evidence-Retrieval, let $\mathcal{G}_q$ denote the gold evidence set and $\mathcal{A}_q$ the answer-supporting evidence. We report three recall metrics:
\begin{equation}
R_{\mathrm{R}}(q)
=
\frac{|\mathcal{C}_q \cap \mathcal{G}_q|}{|\mathcal{G}_q|},
\qquad
R_{\mathrm{FA\text{-}R}}(q)
=
\frac{|\mathcal{C}_q \cap \mathcal{A}_q|}{|\mathcal{A}_q|},
\qquad
R_{\mathrm{TR}}(q)
=
\frac{|\mathcal{P}_q \cap \mathcal{G}_q|}{|\mathcal{G}_q|}.
\end{equation}
Recall measures gold evidence retained in the final set, Final-Answer Recall measures retained evidence directly supporting the answer, and Trajectory Recall measures gold evidence discovered during retrieval. For Answer-Generation QA, we also report Exact Match (EM) and token-level F1.

\begin{figure*}[t]
\centering

\begin{minipage}[t]{0.59\textwidth}
    \vspace{2pt}
    \centering
    \captionof{table}{
      Main results on Answer-Generation QA benchmarks.
      Harness-Search with Qwen3-8B achieves performance comparable
      to similarly sized trained retrieval models, while GPT-OSS-20B
      consistently outperforms existing harness-based methods across
      all three datasets.
    }
    \label{tab:main_table_qa}
    \small
    \setlength{\tabcolsep}{3.5pt}
    \renewcommand{\arraystretch}{1.12}

    \begin{adjustbox}{max width=\linewidth}
    \begin{tabular}{
      @{}
      >{\raggedright\arraybackslash}p{5.5cm}
      *{6}{S[table-format=2.1]}
      @{}
    }
      \toprule
      \multirow{2}{*}{\textbf{Method}}
        & \multicolumn{2}{c}{\textbf{HotpotQA}}
        & \multicolumn{2}{c}{\textbf{2Wiki}}
        & \multicolumn{2}{c}{\textbf{Musique}} \\
      \cmidrule(lr){2-3}
      \cmidrule(lr){4-5}
      \cmidrule(lr){6-7}
        & {EM} & {F1} & {EM} & {F1} & {EM} & {F1} \\
      \midrule

      \multicolumn{7}{l}{\textbf{\textit{A. RAG models with native backbones}}} \\
      Prompt-R1(4B)~\citep{liu2026prompt}      & 44.5 & 52.3 & 48.4 & 54.4 & 18.7 & 26.3 \\
      Search-R1(7B)~\citep{jin2025search}      & 46.5 & 58.0 & 49.6 & 52.6 & 21.2 & 30.0 \\
      KiRAG(7B)~\citep{fang2025kirag}          & 45.1 & 59.8 & 30.7 & 50.6 & 19.2 & 30.0 \\
      StepSearch(7B)~\citep{zheng2025stepsearch} & 38.6 & 50.2 & 36.6 & 43.1 &22.6 & 31.2 \\
      GlobalRAG(7B)~\citep{luo2025globalrag}   & 38.2 & 50.7 & 47.8 & 53.2 & 14.9 & 21.2 \\
      Recomp (Flan-UL2-20B)~\citep{xu2023recomp} & 38.7 & 42.7 & 32.1 & 38.3 & 17.3 & 25.0 \\
      BAVT(20B)~\citep{li2026spend}            & 40.2 & 55.2 & 41.3 & 51.7 & 18.6 & 27.2 \\
      IRCoT(GPT-3.5)~\citep{trivedi2023interleaving} & 45.5 & 58.4 & 35.4 & 45.1 & 19.0 & 30.5 \\
      \textbf{\ourmethod}    & 42.0 & 52.2 & 37.0 & 44.7 & 17.7 & 25.2 \\
      \midrule

      \multicolumn{7}{l}{\textbf{\textit{B. Same-backbone comparison (Except Harness-RL)}}} \\
      Harness-RL(1.5B)~\citep{jiang2026harness} & {\text{--}} & 44.4 & {\text{--}} & 41.6 & {\text{--}} & 27.8 \\
      Harness-RL(3B)~\citep{jiang2026harness}   & {\text{--}} & 57.4 & {\text{--}} & 45.8 & {\text{--}} & 24.3 \\
      ACM~\citep{li2026acm}                     & 47.6 & 55.3 & 33.0 & 41.2 & 22.0 & 30.2 \\
      AgenticRAG~\citep{suresh2026agenticrag}  & 37.0 & 48.2 & 38.9 & 47.3 & 20.3 & 30.3 \\
      Harness-1~\citep{jiang2026harness}       & 41.5 & 48.5 & 23.1 & 37.6 & 18.5 & 27.2 \\
      MemHarness~\citep{wu2026memharness}      & 34.7 & 42.6 & 25.3 & 28.7 & 20.6 & 28.7 \\
      \textbf{\ourmethod} & \textbf{48.5} & \textbf{61.3} & \textbf{45.9} & \textbf{52.9} & \textbf{24.0} & \textbf{31.4} \\
      \bottomrule
    \end{tabular}
    \end{adjustbox}
\end{minipage}
\hfill
\begin{minipage}[t]{0.39\textwidth}
    \vspace{0pt}
    \begin{minipage}[t]{\linewidth}
        \vspace{0pt}
        \centering
        \captionof{table}{
        Table 4: Evidence coverage (\%) across different numbers of interaction turns and token consumption (all agent). Harness-Search continues to improve coverage in later turns.
        }
        \label{tab:harness_turn}
        \footnotesize
        \setlength{\tabcolsep}{3pt}
        \renewcommand{\arraystretch}{1.05}

        \resizebox{\linewidth}{!}{
        \begin{tabular}{lcccccccr}
        \toprule
        \multirow{2}{*}{\textbf{Method}}
        & \multicolumn{5}{c}{\textbf{Turn}} & \multirow{2}{*}{\textbf{\# token}}\\
        \cmidrule(lr){2-6}
        & 1 & 4 & 12 & 20 & 32\\
        \midrule
        s3 & 4.08 & 4.99 & 4.99 & 4.99 & 4.99 & 25.3k \\
        Search-R1 & 17.50 & 20.52 & 20.80 & 20.80 & 20.80 & 35.3k \\
        AgenticRAG & 17.33 & 29.80 & 29.80 & 29.80 & 29.80& 83.1k\\
        
        MemHarness & 23.76 & 28.13 & 30.19 & 30.19 & 30.19 & 113k\\
        Harness-1 & 0 & 1.22 & 35.75 & 47.63 & 56.25 & 230k \\
        \hdashline
        \textbf{\ourmethod} & 0 & 2.38 & 39.79 & 56.25 & 65.57 & 260k\\
        \bottomrule
        \end{tabular}
        }
        
    \end{minipage}
    
    \vspace{0.8em}
    
    \begin{minipage}[t]{\linewidth}
        \vspace{0pt}
        \centering
        \captionof{table}{Effect of Harness-Search-guided SFT. Fine-tuning on teacher trajectories improves all metrics across benchmarks, compared with the untuned model.
        }
        \label{tab:harness_sft}
        \footnotesize
        \setlength{\tabcolsep}{3pt}
        \renewcommand{\arraystretch}{1.05}

        \resizebox{\linewidth}{!}{
        \begin{tabular}{llcccc}
        \toprule
        \textbf{Method}
        & \textbf{Metrics}
        & \textbf{BC+}
        & \textbf{Web}
        & \textbf{Sec}
        & \textbf{LongSeal}\\
        \midrule

        \multirow{3}{*}{GPT-OSS-20B}
        & R    & 57.81 & 62.19 & 37.97 & 88.98\\
        & FA-R & 61.51 & 77.98 & 45.44 & 88.98\\
        & T-R  & 66.86 & 70.03 & 44.65 & 94.88\\
        \hdashline


        \multirow{3}{*}{Harness-Search (SFT)}
        & R    & 65.12 & 75.34 & 46.57 & 90.21\\
        & FA-R & 70.45 & 80.18 & 50.63 & 90.21\\
        & T-R  & 73.29 & 81.52 & 59.11 & 97.36\\

        \bottomrule
        \end{tabular}
        }
    \end{minipage}
    
\end{minipage}

\end{figure*}

\textbf{Implementation Details.}
All harness baselines use GPT-OSS-20B as the policy model, enabling a controlled comparison with \ourmethod under the same policy backbone. Unless otherwise specified, we set the maximum number of interaction turns to $T_{\max}=40$, the maximum number of consecutive search operations to $S_{\max}=5$, and the evidence capacity to $B=60$. The Retrieval Policy, Memory Operator, and Summary Auditor use the same backbone model. For Answer-Generation QA, answers are generated from the curated evidence set using the same model as the retrieval policy.

\subsection{Main Results}
\textbf{Evidence retrieval.}
As shown in Table~\ref{tab:recall-results}, with GPT-OSS-20B as the policy model, \ourmethod consistently outperforms all harness baselines in final evidence quality across the four benchmarks. Compared with the strongest harness baseline on each dataset, it improves Recall by 4.60--27.92 points and Final-Answer Recall by 12.34--30.13 points. \ourmethod also improves Trajectory Recall on three of the four benchmarks by 10.50--23.56 points, while remaining comparable on Web Synthetic (70.03 vs.\ 70.92). Beyond harness-based methods, \ourmethod also outperforms all compared RAG models on R, FA-R, and TR across all four benchmarks. The same harness can be directly instantiated with different model families, including Qwen, Gemini, and GPT-series models.

\textbf{Answer generation.}
Table~\ref{tab:main_table_qa} leads to two main observations. First, even with Qwen3-8B as the policy model, \ourmethod achieves performance comparable to similarly sized retrieval models that require task-specific training, indicating that the proposed harness can elicit competitive multi-hop QA performance without relying on additional policy training. Second, when instantiated with GPT-OSS-20B, \ourmethod consistently outperforms all compared harness-based methods across the reported metrics on all three datasets.

\begin{figure*}[t]
\centering

\begin{minipage}[t]{0.54\textwidth}
    \centering
    \small
    \captionof{table}{
    Results for component ablations and single-agent alternatives.
    $\Delta R$ is the paired Recall difference between the full system and each
    control ($\Delta R=R_{\mathrm{full}}-R_{\mathrm{ablation}}$), positive
    values favor the full system.
    }
    \label{tab:agent_ablation}

    \resizebox{\linewidth}{!}{%
    \begin{tabular}{lcc}
    \toprule
    \textbf{System or intervention} & \textbf{Recall} & \textbf{$\Delta R$ (95\% CI)} \\
    \midrule
    \multicolumn{3}{l}{\textit{Component ablations: which functions matter?}} \\
    Full Harness-Search system
        & \textbf{57.81} & --- \\
    \quad -- Memory Operator
        & 44.74 & $+13.07\ [10.71,15.43]$ \\
    \quad -- Summary Auditor
        & 52.07 & $+5.74\ [3.52,7.90]$ \\
    \quad -- Relevance Judge
        & 52.98 & $+4.83\ [2.78,6.86]$ \\
    \quad -- Direction Planner
        & 52.58 & $+5.23\ [3.52,6.94]$ \\
    \midrule
    \multicolumn{3}{l}{\textit{Monolithic controls: can a single agent recover the gain?}} \\
    Single agent with structured tools
        & 54.40 & $+3.41\ [1.29,5.55]$ \\
    Single agent with three-role prompting
        & 55.35 & $+2.46\ [0.33,4.55]$ \\
    \bottomrule
    \end{tabular}%
    }
\end{minipage}
\hfill
\begin{minipage}[t]{0.45\textwidth}
    \vspace{0pt}
    \centering
    \includegraphics[width=0.72\linewidth]{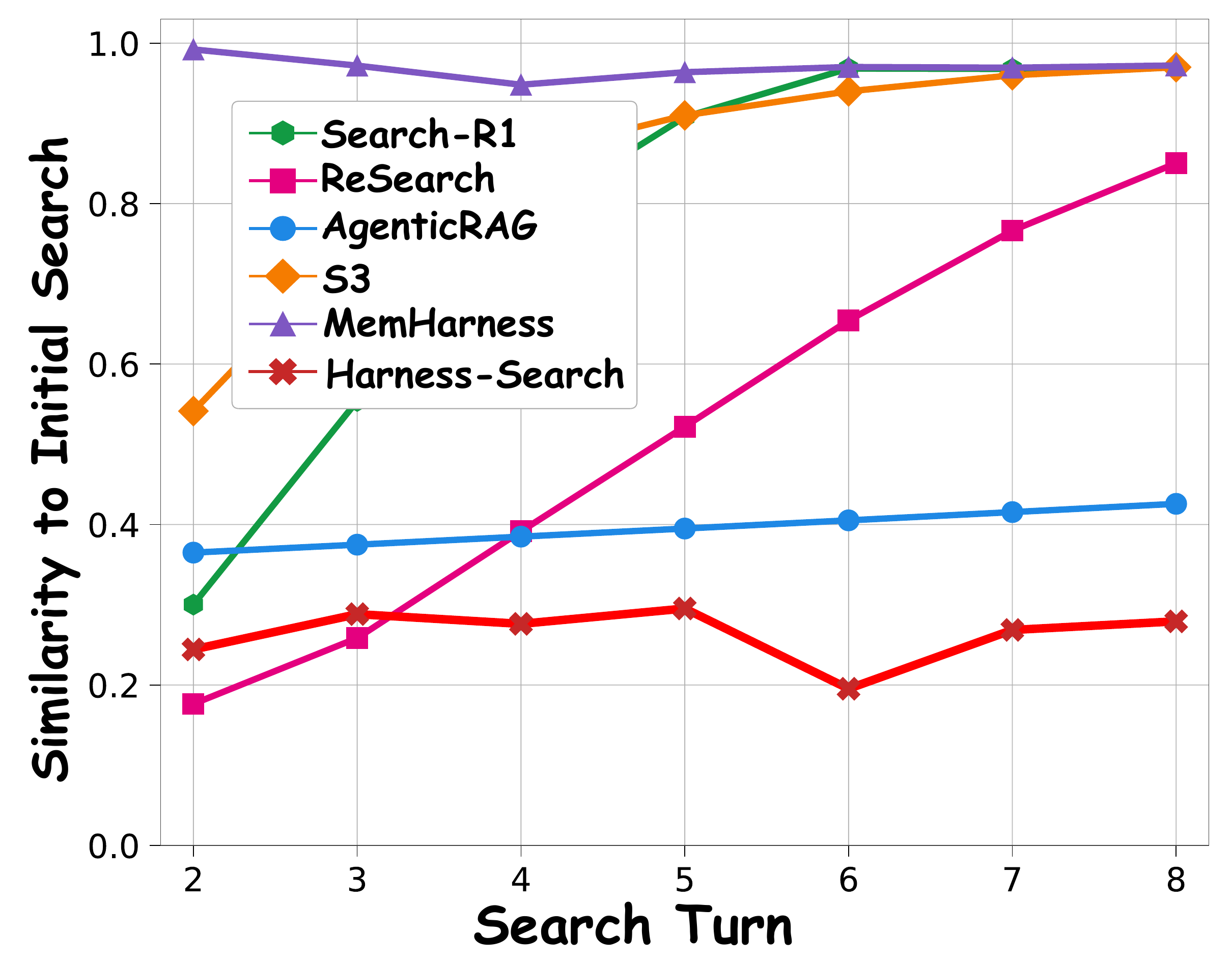}
    \caption{
    Search-term similarity to the first search turn. Higher values indicate greater repetition. \ourmethod maintains low, relatively stable similarity throughout.
    }
    \label{fig:search_similarity}
\end{minipage}

\end{figure*}

%% file: section/5_analysis.tex
\subsection{What Drives the Gains of Harness-Search?}

\textbf{Which components matter?}
To investigate each specialized component, we compare the full system with variants that individually remove the Memory Operator, Summary Auditor, or Document Relevance Judge. These ablations measure the contribution of each function within the complete architecture, rather than the standalone capability of an individual component.

In Table~\ref{tab:agent_ablation}, removing the Memory Operator produces the largest observed degradation: Recall decreases from 57.81 to 44.74, with a paired difference of 13.07 points. Removing the Summary Auditor reduces Recall to 52.07, corresponding to drops of 5.74 points. The particularly large effect of removing the Memory Operator shows that retrieving potentially useful documents is insufficient if the evidence is not effectively maintained and consolidated for subsequent decisions. The other ablations indicate  document-level relevance assessment and termination auditing also provide benefits that the remaining components do not fully compensate for. 

\textbf{Can a monolithic agent recover the gains?}
The component ablations establish that the specialized functions are useful, but do not establish that their organization into separate roles is necessary. A single agent might recover the same benefits through better tools, more explicit instructions, or additional opportunities to interact with the environment. We therefore evaluate two monolithic alternatives, each targeting a different explanation: structured tools test whether access to the relevant operations is sufficient and three-role prompting tests whether describing the specialized responsibilities within one agent is sufficient. None of these alternatives matches the full system. The single agent with structured tools achieves 54.40 Recall, remaining 3.41 points below Harness-Search. Three-role prompting achieves 55.35 Recall at the displayed precision, with paired deficits of 2.46 points. These comparisons distinguish making an operation available from organizing its use throughout the retrieval process.

\subsection{How Does Harness-Search Alter Retrieval Trajectories?}

\textbf{Search diversification.}
We first investigate whether \ourmethod moves beyond the parameters of the initial search. The improvement in recall could stem either from making better selections among documents retrieved for similar queries or from a shift in the search direction that uncovers more relevant documents. To examine the latter possibility, Figure~\ref{fig:search_similarity} tracks the similarity between the search terms used in each round and the vocabulary employed in the initial search.

\ourmethod maintains consistently low similarity to the initial search across later turns, whereas several baselines increasingly return to terms resembling their initial queries. This pattern is consistent with reduced anchoring to the initial formulation and a greater tendency to redirect exploration. Such redirection provides opportunities to recover evidence that the initial search direction may have missed, rather than repeatedly relying on closely related formulations.

\begin{wrapfigure}[14]{r}{0.3\textwidth}
    \centering
    \includegraphics[width=0.96\linewidth]{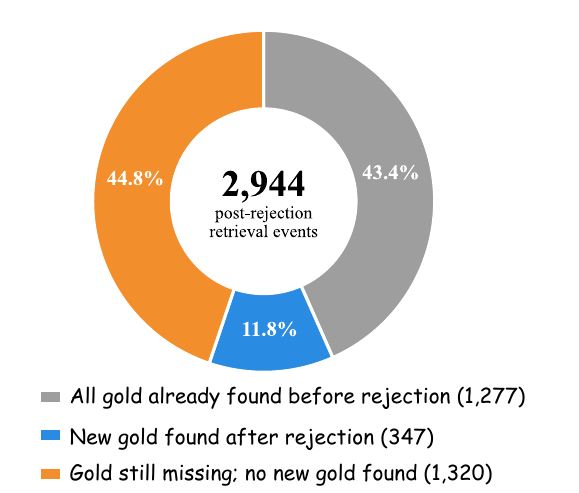}
    \caption{Behavior following a Summary Auditor rejection.}
    \label{fig:auditor_behavior}
\end{wrapfigure}
\textbf{Recovery after failed termination.}
To examine how the Summary Auditor contributes to Recall, we measure how often post-rejection searches recover previously unretrieved gold documents in $\mathcal{A}_q$, the evidence set directly supporting the final answer. Figure~\ref{fig:auditor_behavior} shows that all gold documents had already been retrieved before rejection in 1,277 events. Among 1,667 events with missing gold documents, subsequent search recovered at least one new gold document in 347 cases (about one-fifth), while 1,320 yielded none. Together with the 5.74-point Recall drop without the auditor (Table~\ref{tab:agent_ablation}), these results support evidence recovery through continued search as one mechanism behind its benefit. However, the substantial fraction of events without recovery indicates that rejection creates an opportunity to fill evidence gaps rather than guaranteeing their resolution. More settings and analyses are presented in Appendix~\ref{app:sft}.

\subsection{Supervised Fine-Tuning}
The preceding experiments evaluate Harness-Search with off-the-shelf models without task-specific training. Prior work has shown that trajectory-based SFT can improve tool-using agents~\citep{qin2023toolllmfacilitatinglargelanguage,chen2023fireactlanguageagentfinetuning}. We therefore adopt this approach to fine-tune our model using supervision selected by Harness-Search. We use GPT-5.5 trajectories collected on SEC training queries and perform full-parameter SFT of GPT-OSS-20B for three epochs using ms-swift~\citep{zhao2024swiftascalablelightweightinfrastructure}. The training set contains 1,132 responses including 610 Policy examples, 299 Memory examples, and 223 Auditor examples. In Table~\ref{tab:harness_sft}, Harness-Search (SFT) improves all three metrics across all four benchmarks relative to the untuned model. Recall increases by 7.31, 13.15, 8.60, and 1.23 points on BC+, Web, SEC, and LongSeal, respectively. Using SFT alone, it also outperforms Harness-1 (SFT+RL) on every reported metric. These results suggest that Harness-Search provides useful training supervision in addition to its inference-time benefits.

\subsection{Additional Analyses and Experiments}

As shown in Table~\ref{tab:harness_turn} and the bottom-right panel of Figure~\ref{fig:intro_figure}, Harness-Search progressively improves evidence coverage as the number of interaction turns increases, with gains continuing into later turns. Compared to the best baseline, coverage improved by 16.6\% while the total token count increased by only 13.0\%. Besides, Appendix~\ref{app:analysis} presents additional experiments and analyses. We examine how the performance of \ourmethod{} depends on the interaction budget and the frequency of evidence curation (Appendix~\ref{app:budget} and Figure~\ref{fig:budget_sensitivity}). Our choice of $T_{\max}=40$ and $S_{\max}=5$ offers a favorable quality--latency trade-off. To assess whether the decomposed roles have different model requirements, we keep the Retrieve Policy fixed and independently vary the models used by the Memory Operator and Summary Auditor (Appendix~\ref{app:role} and Figure~\ref{fig:heatmap}). Performance is sensitive to auxiliary model choices, with Qwen3-32B in both roles yielding 66.60\% Recall.

%% file: section/appendix.tex
\section{Algorithm}
\label{app:algorithm}

\begin{algorithm}[]
\caption{\ourmethod{}}
\label{alg:harness_search}
\small
\begin{algorithmic}[1]
\REQUIRE Query $q$, corpus $\mathcal{D}$, initial direction $I_0$,
         evidence capacity $B$, and execution budgets $T_{\max}, S_{\max}$
\STATE $\mathcal{A} \gets
       \{\texttt{search},\texttt{read},\texttt{redirect},
         \texttt{curate},\texttt{end}\}$
\STATE $s_0 \gets 0$

\FOR{$t = 0,\ldots,T_{\max}-1$}

    \IF{$s_t \geq S_{\max}$}
        \STATE $\mathcal{A}_t \gets \{\texttt{curate}\}$
    \ELSIF{$a_{t-1} = \texttt{curate}$}
        \STATE $\mathcal{A}_t \gets
               \{\texttt{redirect},\texttt{end}\}$
    \ELSE
        \STATE $\mathcal{A}_t \gets \mathcal{A}$
    \ENDIF

    \STATE $(a_t,u_t) \gets
           \pi_{\mathrm{ret}}(\mathcal{M}_t,\mathcal{A}_t)$
           \COMMENT{$u_t$ is the reasoning}

    \IF{$a_t \in \{\texttt{search},\texttt{read}\}$}
        \STATE $\mathcal{M}_{t+1} \gets
               \mathrm{Mem.RelevanceJudge}(\mathcal{M}_t)$
        \IF{$a_t = \texttt{search}$}
            \STATE $s_{t+1} \gets s_t + 1$
        \ENDIF

    \ELSIF{$a_t = \texttt{redirect}$}
        \STATE $(\mathcal{I}_{t+1}, \mathcal{L}_{t+1}) \gets
               \mathrm{Mem.DirectionPlanner}(\mathcal{M}_t,u_t)$

    \ELSIF{$a_t = \texttt{curate}$}
        \STATE $\mathcal{M}_{t+1} \gets
               \mathrm{Mem.Curate}(\mathcal{M}_t,u_t,B)$
        \STATE $s_{t+1} \gets 0$

    \ELSIF{$a_t = \texttt{end}$}
        \STATE $(y_t,v_t,g_t) \gets
               \pi_{\mathrm{audit}}(q,\mathcal{C}_t)$
               \COMMENT{$g_t$ is the reasoning}
        \IF{$v_t = \texttt{answer\_ready}$}
            \STATE \textbf{return} $\mathcal{C}_t$
        \ELSE
            \STATE $\mathcal{A}_{t+1} \gets \{ \texttt{redirect} \}, s_0 \gets 0$
        \ENDIF
    \ENDIF

\ENDFOR

\STATE \textbf{return} $\mathcal{C}_{t}$
\end{algorithmic}
\end{algorithm}

\section{Experiment Details}
\label{app:experiment}
\subsection{Model Usage and License}
\label{app:model}
In this paper, we use several types of models. We will release their versions and link and corresponding license in this section.

\paragraph{Small Scale Model} 
\begin{itemize}
    \item[] Qwen3-8B:
    \begin{itemize}
        \item Source: \href{https://huggingface.co/Qwen/Qwen3-8B}{https://huggingface.co/Qwen/Qwen3-8B}
        \item License: Apache 2.0
    \end{itemize}
    \item[] gpt-oss-20b: 
    \begin{itemize}
        \item Source:\href{https://huggingface.co/openai/gpt-oss-20b}{https://huggingface.co/openai/gpt-oss-20b}
        \item License: Apache 2.0
    \end{itemize}
    \item[] Qwen3.5-27B:
    \begin{itemize}
        \item Source: \href{https://huggingface.co/Qwen/Qwen3.5-27B}{https://huggingface.co/Qwen/Qwen3.5-27B}
        \item License: Apache 2.0
    \end{itemize}
    \item[] Qwen3.8-Max:
    \begin{itemize}
        \item Source: \href{https://huggingface.co/Qwen/Qwen3.8-27B}{https://huggingface.co/Qwen/Qwen3.8-2.4T-A95B} 
        \item License: Apache 2.0
    \end{itemize}
    
    \item[] Harness-1(20B): 
    \begin{itemize}
        \item Source: 
        \href{https://huggingface.co/pat-jj/harness-1}{https://huggingface.co/pat-jj/harness-1}
    \item License: Apache 2.0
    \end{itemize}
    
    \item[] Search-R1(32B): 
    \begin{itemize}
        \item Source: \href{https://huggingface.co/PeterJinGo/SearchR1-nq_hotpotqa_train-qwen2.5-32b-em-grpo-v0.3}{https://huggingface.co/PeterJinGo/SearchR1-nq\_hotpotqa\_train-qwen2.5-32b-em-grpo-v0.3}
        \item License: Apache 2.0
    \end{itemize}
    \item[] Tongyi DeepResearch(30B): 
    \begin{itemize}
        \item Source: \href{https://huggingface.co/Alibaba-NLP/Tongyi-DeepResearch-30B-A3B}{https://huggingface.co/Alibaba-NLP/Tongyi-DeepResearch-30B-A3B}
        \item License: Apache 2.0
    \end{itemize}
    \item[] s3(7B): 
    \begin{itemize}
        \item Source: \href{https://huggingface.co/pat-jj/s3-8-3-3-20steps}{https://huggingface.co/pat-jj/s3-8-3-3-20steps}
        \item License: Apache 2.0
    \end{itemize}
    \item[] ReSearch(32B): 
    \begin{itemize}
        \item Source: \href{https://huggingface.co/agentrl/ReSearch-Qwen-32B}{https://huggingface.co/agentrl/ReSearch-Qwen-32B}
        \item License: MIT License
    \end{itemize}
\end{itemize}

\paragraph{Large Scale Model} 
\begin{itemize}
    \item[] ChatGPT-5.6-terra: \href{https://developers.openai.com/api/docs/models/gpt-5.6-terra}{https://developers.openai.com/api/docs/models/gpt-5.6-terra}
    \item[] Gemini 3.1 pro preview: \href{https://deepmind.google/models/gemini/pro/}{https://deepmind.google/models/gemini/pro/}
\end{itemize}

\subsection{Retrieval backend.}
\label{app:retrieve}
For each search query, we retrieve candidate documents using BM25~\citep{robertson2009probabilistic} and Qwen3-Embedding-8B~\citep{qwen3embedding} to capture lexical and semantic relevance, respectively. We combine the resulting ranked lists using reciprocal rank fusion (RRF)~\citep{cormack2009reciprocal} and then rerank the fused candidates with Qwen3-Reranker-8B.

\subsection{Evaluation Datasets}
\label{app:datasets}
We evaluate on seven benchmarks:
\begin{itemize}
    \item \textbf{BrowseComp-Plus}~\citep{chen2025browsecomp}:
    830 queries from \texttt{Tevatron/browsecomp-plus}, paired with
    100,195 corpus chunks from \texttt{Tevatron/browsecomp-plus-corpus}.

    \item \textbf{Web}~\citep{bashir2026context1}:
    554 queries from \texttt{kellyhongg/web\_1\_17\_test}, with
    54,735 chunks from \texttt{pat-jj/harness-1-train-data}
    (\texttt{corpora/web/test}).

    \item \textbf{SEC}~\citep{bashir2026context1}:
    691 queries from \texttt{kellyhongg/sec\_test\_new}, evaluated against
    2,115,106 chunks from \texttt{corpora/sec/train}.

    \item \textbf{LongSealQA}~\citep{pham2026sealqa}:
    254 queries from \texttt{vtllms/sealqa} (\texttt{longseal/test}),
    with 6,283$^\dagger$ chunks from the \texttt{30\_docs} corpus.
    
    \item \textbf{HotpotQA}~\citep{yang2018hotpotqa}: 7,405 compositional questions from the development split of \texttt{hotpotqa/hotpot\_qa}. We use the full-corpus retrieval setting with the shared Wikipedia collection \texttt{wiki18\_100w}.
    \item\textbf{2WikiMultiHopQA}~\citep{ho2020constructing}:
    12,576 multi-hop questions from the development split of \texttt{Alab-NII/2wikimultihop}, evaluated against the shared Wikipedia corpus \texttt{wiki18\_100w}, containing approximately one million passages.

    \item \textbf{MuSiQue}~\citep{trivedi2022musique}:
    2,417 multi-hop questions from the development split of \texttt{stonybrooknlp/musique}, paired with the same approximately one-million-passage
    Wikipedia corpus, \texttt{wiki18\_100w}.
\end{itemize}

\noindent
$^\dagger$ LongSealQA does not use a single global corpus. Each query is
associated with its own document bundle from the \texttt{30\_docs} field.
The test set contains 6,283 query-document instances in total
(6,250 unique document URLs).

\subsection{Tools}
\label{app:tools}
The tools used in our experiments are summarized in Table~\ref{tab:tool_signatures}. In particular, \texttt{fan\_out\_search(queries)}, \texttt{search(query)}, and \texttt{grep\_corpus(pattern)} are categorized as search operations.
\begin{table*}[htbp]
    \centering
    \small
    \setlength{\tabcolsep}{8pt}
    \renewcommand{\arraystretch}{1.08}
    \begin{tabular}{@{}p{0.36\textwidth}p{0.59\textwidth}@{}}
        \toprule
        \textbf{Tool signature} & \textbf{Effect and output} \\
        \midrule

        \texttt{fan\_out\_search(queries)}
        &
        Executes up to five complementary hybrid searches and returns the
        combined ranked document IDs and text snippets.
        \\

        \texttt{search(query)}
        &
        Performs a single hybrid search using sparse and dense retrieval,
        followed by fusion and reranking; returns ranked document IDs and
        text snippets.
        \\

        \texttt{grep\_corpus(pattern)}
        &
        Matches an exact or regular-expression pattern against the corpus
        and returns matching document IDs and text excerpts.
        \\
        \hdashline

        \texttt{read(doc\_id)}
        &
        Retrieves the full content of the specified document and adds it to the document memory.
        \\
        \texttt{review\_docs(doc\_ids)}
        &
        Re-renders up to five previously retrieved documents from outer memory without issuing an additional corpus call.
        \\
        \hdashline

        \texttt{redirect(doc\_ids, reasoning)}
        &
        Sends up to five newly retrieved candidate documents and the policy's
        rationale to the intent model; returns an updated active retrieval
        intent.
        \\
        \hdashline

        \texttt{curate(add\_ids, remove\_ids)}
        &
        Updates the curated evidence set with documents from the candidate
        pool and returns the resulting set together with capacity feedback.
        \\
        \hdashline

        \texttt{end(reasoning)}
        &
        Terminates the retrieval episode and returns the curated evidence set
        as the final retrieval output.
        \\

        \bottomrule
    \end{tabular}
    \caption{Intent Curator tool signatures.}
    \label{tab:tool_signatures}
\end{table*}

\subsection{Compared Baselines}
\label{app:baselines}

\paragraph{Search-R1.}
We reproduced Search-R1 using the publicly released \texttt{PeterJinGo/SearchR1-nq\_hotpotqa\_train-qwen2.5-32b-em-grpo-v0.3} checkpoint. We implemented the inference loop according to the model's native interaction protocol: at each turn, the model emitted either a \texttt{<search>...</search>} action or a final \texttt{<answer>...</answer>}. Search actions were executed against exactly the same corpus and retrieval indexes used by our method.  We allowed at most 40 search turns, generated at most 1,024 tokens per turn, and used greedy decoding ($T=0$). Because Search-R1 does not explicitly maintain a separate evidence pool, its curated set was obtained from the unique documents returned across all search turns, ordered by first appearance and within-query retrieval rank, and truncated to 60 documents. The complete trajectory, including every query, retrieved document identifier, and final answer, was saved.

\paragraph{Tongyi DeepResearch.}
We reproduced Tongyi DeepResearch with the released \texttt{Tongyi-DeepResearch-30B-A3B} checkpoint. We retained the model's native deep-research agent loop and prompting format, where actions are represented as JSON objects inside \texttt{<tool\_call>...</tool\_call>} tags and observations are returned inside \texttt{<tool\_response>...</tool\_response>} tags. The original Web-oriented \texttt{search} and \texttt{visit} operations were adapted only at the tool boundary to access our fixed benchmark corpora: \texttt{search} queried the same BM25--Qdrant hybrid indexes and \texttt{Qwen3-Embedding-8B} embeddings used by our method, while \texttt{visit} returned the full text of a selected corpus document. LongSealQA instead used its query-specific local document collection. Each episode was allowed up to 40 tool calls, with a maximum of 4,096 newly generated tokens per turn, temperature $0.6$, nucleus-sampling parameter $p=0.95$, and presence penalty $1.1$. For curated-set evaluation, documents explicitly accessed through \texttt{visit} were placed first, followed by additional unique documents surfaced by \texttt{search} in order of first appearance and retrieval rank, up to the common limit of 60 documents.

\paragraph{S3.}
We reproduced S3 using its publicly released \texttt{s3-8-3-3-20steps} checkpoint and native iterative retrieval protocol. Following the released inference configuration, the agent performs at most three search rounds; in each round, it retrieves the top eight documents and selects up to three documents to retain as evidence for subsequent reasoning. We served the checkpoint locally with vLLM and replaced only the original retrieval environment with our benchmark-specific search interface. The underlying corpora, document identifiers, dataset splits, and retrieval behavior were kept identical to those used in our main experiments.

\paragraph{ReSearch.}
We reproduced ReSearch using the publicly released \texttt{agentrl/ReSearch-Qwen-32B} checkpoint and its native interleaved \texttt{<think>}, \texttt{<search>}, \texttt{<result>}, and \texttt{<answer>} protocol. Following the official evaluation setting, we used deterministic decoding, returned the top five documents for each search request, allowed up to 32 search rounds, and imposed an 8,192-token generation budget shared across the complete interaction trajectory. We retained the original ReSearch policy and prompting format, modifying only the environment adapter so that every search action queried the same benchmark-specific corpus and hybrid retrieval infrastructure as our main method. ReSearch does not maintain a separate persistent evidence set, so we constructed its curated set from the unique documents appearing in \texttt{<result>} observations across the trajectory, preserving first-appearance order and retrieval rank and retaining at most 60 documents. 

\paragraph{AgenticRAG.}
We implemented AgenticRAG using \texttt{gpt-oss-20b} as the backbone model. The agent was prompted to iteratively decide whether additional evidence was required, formulate or revise search queries, inspect retrieved documents, and produce a final evidence-grounded answer. We preserved this search--reason--reflect interaction pattern. We then saved all model decisions, tool calls, retrieved document identifiers, intermediate reasoning states, and final answers for evaluation.

\paragraph{ACM.} We reproduced Agentic Context Management (ACM) using \texttt{gpt-oss-20b} as the base model and retained its explicit context-management mechanism. In addition to the standard \texttt{search} and \texttt{get\_document} operations, the agent could invoke \texttt{manage\_context} to replace an increasingly long interaction segment with a compact summary. The complete pre-compression segment was archived losslessly and assigned a persistent summary identifier; the agent could subsequently call \texttt{query\_memory} with this identifier to recover specific facts, document identifiers, unresolved leads, or earlier observations from the archived context. Context compression therefore reduced the active prompt length without permanently discarding information. We adapted only the retrieval and evaluation interfaces so that searches accessed the same fixed corpora, document identifiers, and retrieval indexes as our method, with LongSealQA using its query-specific local collection. Each episode was allowed up to 40 tool calls, and we saved the complete active trajectory, compressed summaries, lossless memory archives, memory-access operations, retrieved documents, and final answer. 

\paragraph{Harness-1.}
We evaluated Harness-1 with the base \texttt{gpt-oss-20b} model. We retained its native agentic inference loop, in which the model alternates between reasoning and corpus interaction through search and document-reading tools until it emits a final response or exhausts the interaction budget. We modified only the retrieval adapter: tool calls were executed against the same fixed benchmark corpora and indexes used by our method, and the model was never given access to gold document identifiers or evaluation annotations. 

\paragraph{MemHarness.}
We reproduced MemHarness using its released memory-augmented inference design with \texttt{gpt-oss-20b} as the base model. Following the original setting, memory was maintained at the dataset level rather than constructed independently for each test query. For a new query, the system retrieved semantically similar training queries together with their stored search trajectories and exposed the selected experiences as demonstrations or reusable research memory. The backbone model then conducted the current episode using its own reasoning and tool calls, conditioned on the retrieved memories. We reused the released memory-management logic and adapted only the dataset, retrieval, and evaluation interfaces: current-query searches accessed exactly the same fixed corpora and retrieval indexes as our method, while memory retrieval was restricted to previously available training trajectories to prevent test-set leakage. The retrieved memories, newly executed searches, document identifiers, and final response were all saved for evaluation.

\section{Reproducibility Statement}
\label{app:reproducibility}


\paragraph{Hardware Environment.}
We conduct all experiments on a single compute node equipped with $8\times$ NVIDIA H100 GPUs, each with 80\,GB of memory. The policy model, dense retriever, and reranker are served using
tensor parallelism across the eight GPUs.
Unless otherwise specified, all methods are evaluated on the same
hardware configuration.

\paragraph{Software Environment.} 
All experiments were conducted on Ubuntu 22.04.5 LTS. The software environment was based on Python 3.11. Model implementation and training were performed using PyTorch 2.6.0 with CUDA support, together with HuggingFace Transformers 4.57.6. Unless otherwise specified, all models and libraries were used with their
default parameter settings.

\paragraph{Inference Configuration.}
Each retrieval episode is limited to 40 policy turns, and the policy generates at most 2,048 tokens per turn. The complete model context is limited to 32,768 tokens. Our method permits at most five candidate-discovery calls within each intent stage before triggering an evidence-curation checkpoint. The working memory preserves the active intent, candidate pool, curated evidence set, and search history, while the five most recent action--observation pairs are retained in full. We use a policy sampling temperature of 1.0, an intent-revision temperature of 0.2, and deterministic decoding for the reference and summary judges. The policy, reranker, and embedding services share the same eight GPUs, with GPU-memory-utilization limits of 0.55, 0.22, and 0.18, respectively.

\section{More Analysis on Ablation Study and Discussion}
\label{app:analysis}

\subsection{Repeated Experiments Across Same / Random Seeds}
\label{app:repeat}

\begin{table}

\centering
\caption{
Repeated experiments across same and random seeds
}
\label{tab:repeat}
\small
\begin{tabular}{lcccc}
\toprule
\textbf{Seed} & \textbf{BC+} & \textbf{Web} & \textbf{Sec} & \textbf{LongSeal} \\
\midrule
20  & 55.00 & 59.20 & 35.20 & 86.00 \\
42  & 57.81 & 62.19 & 37.97 & 88.98 \\
314 & 61.50 & 65.50 & 41.00 & 92.20 \\
\textbf{Avg.} &
$58.10{\color{gray}\scriptstyle \pm 3.26}$ &
$62.30{\color{gray}\scriptstyle \pm 3.15}$ &
$38.06{\color{gray}\scriptstyle \pm 2.90}$ &
$89.06{\color{gray}\scriptstyle \pm 3.10}$ \\
\midrule
42 & 56.32 & 61.05 & 36.41 & 87.52 \\
42 & 58.27 & 62.48 & 37.83 & 89.17 \\
42 & 58.84 & 63.04 & 39.67 & 90.25 \\
\textbf{Avg.} & $57.81{\color{gray}\scriptstyle \pm 1.32}$ & $62.19{\color{gray}\scriptstyle \pm 1.03}$ & $37.97{\color{gray}\scriptstyle \pm 1.63}$ & $88.98{\color{gray}\scriptstyle \pm 1.37}$ \\
\bottomrule
\end{tabular}
\end{table}

Our method involves long-horizon, multi-turn interactions, where small stochastic differences can accumulate and lead to different trajectories and final results even under identical settings. We therefore repeat the main experiment under both different random seeds(Table~\ref{tab:repeat} Upper) and the same seed(Table~\ref{tab:repeat} Lower) to assess run-to-run reproducibility and sensitivity to generation randomness. All other experimental settings, including the test set, model configuration, retrieval services, and turn budget, are kept unchanged.

\subsection{Multi-Agent Decomposition}
\label{app:multiagent}

We conduct a controlled ablation study to determine whether the gains of our method arise from its three functional roles or can be reproduced by a monolithic agent with equivalent prompts, tools, or computation. All conditions are evaluated on the complete BrowseComp-Plus test split of 830 queries using GPT-OSS-20B, temperature $0.2$, top-$p$ $0.9$, a default 40-turn limit, and identical retrieval and reranking services. We report their paired difference $\Delta R=R_{\mathrm{full}}-R_{\mathrm{ablation}}$. Confidence intervals are obtained from 10,000 paired bootstrap samples. The complete experimental results are presented in Table~\ref{tab:agent_ablation_full}.

\begin{table*}[ht]
\centering
\caption{
Full Multi-agent decomposition study. $R_{\mathrm{ctrl}}$ is the Recall of each control. $\Delta R=R_{\mathrm{full}}-R_{\mathrm{ablation}}$; positive values favor the complete system. FA-R denotes final-answer Recall, TR denotes trajectory Recall. All Recall values are percentages.
}
\small
\setlength{\tabcolsep}{4.2pt}
\label{tab:agent_ablation_full}
\resizebox{\textwidth}{!}{
\begin{tabular}{llrrrrrrr}
\toprule
Category & System or intervention
& $R_{\mathrm{ablation}}$
& $\Delta R$ (95\% CI)
& FA-R
& TR \\
\midrule

Full
& --
& \textbf{57.81}
& ---
& 61.51
& 66.86 \\

\midrule
\multirow{4}{*}{Component}
& Without Memory Operator
& 44.74
& $+13.07\ [10.71,15.43]$
& 49.89
& 59.08 \\

& Without Summary Auditor
& 52.07
& $+5.74\ [3.52,7.90]$
& 56.67
& 65.34 \\

& Without Document Relevance Judge
& 52.98
& $+4.83\ [2.78,6.86]$
& 55.99
& 64.26 \\

& Without Direction Planner
& 52.58
& $+5.23\ [3.52,6.94]$
& 54.43
& 63.12 \\

\midrule
\multirow{2}{*}{Monolithic}

& True single agent + structured tools
& 54.40
& $+3.41\ [1.29,5.55]$
& 55.93
& 63.34 \\

& True single agent + three-role prompt
& 55.35
& $+2.46\ [0.33,4.55]$
& 58.25
& 64.39 \\


\bottomrule
\end{tabular}
}
\end{table*}

\subsection{Sensitivity to Interaction and Search Budgets}
\label{app:budget}

\begin{figure*}[b!]
    \centering
    \begin{subfigure}{0.46\textwidth}
        \centering
        \includegraphics[width=\linewidth]{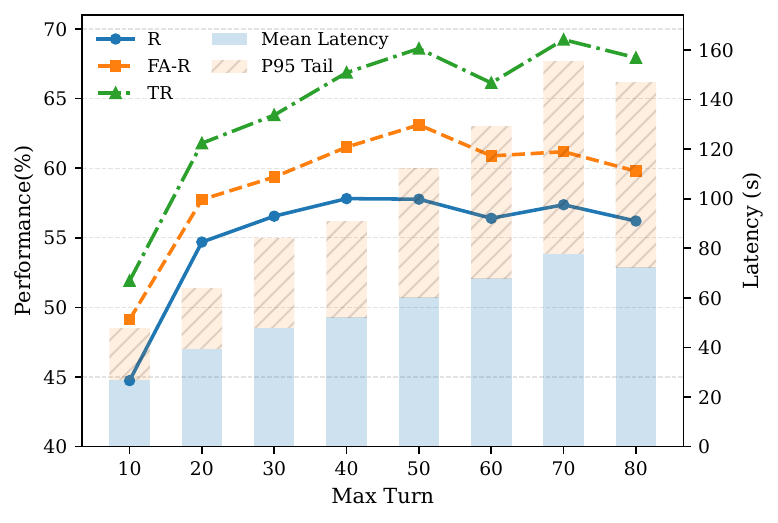}
        \caption{Max Turn sensitivity.}
    \end{subfigure}
    \begin{subfigure}{0.46\textwidth}
        \centering
        \includegraphics[width=\linewidth]{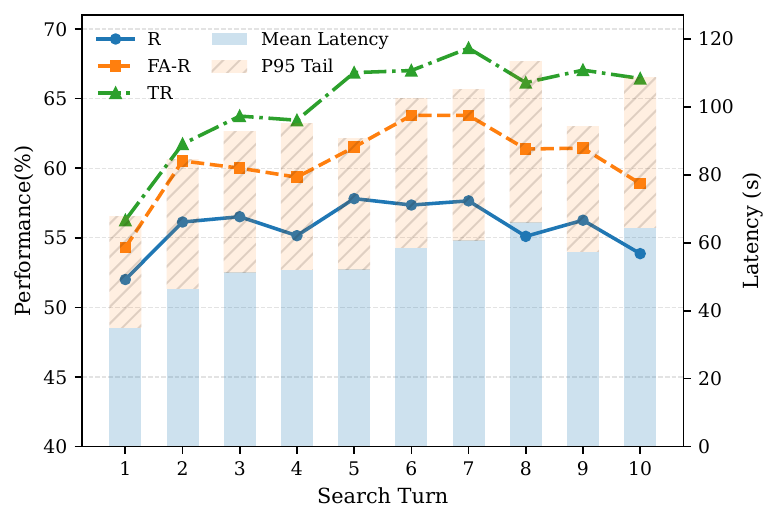}
        \caption{Search Turn sensitivity.}
    \end{subfigure}
    \caption{
    Sensitivity to inference-time interaction.
    }
 \label{fig:budget_sensitivity}
\end{figure*}

We evaluate the sensitivity of our method to two inference-time budget parameters: the maximum total number of agent interaction turns, denoted by $T_{\max}$, and the maximum number of consecutive \texttt{search} actions allowed before a \texttt{curate} operation is required, denoted by $S_{\max}$. The former determines the maximum trajectory length, while the latter determines how often newly retrieved evidence must be consolidated into working memory. When no explicit limit on consecutive \texttt{search} actions is imposed (search turn = no limitation), we instead allow the model to decide autonomously when to invoke \texttt{curate}; in practice, the model may fail to curate in a timely manner, leading to degraded recall and continued searching until the overall turn budget is exhausted. We vary $T_{\max}$ from 10 to 80 with $S_{\max}=5$, and vary $S_{\max}$ from 1 to 10 with $T_{\max}=40$. All other inference settings are kept fixed, with GPT-OSS-20B as the base model and BrowseComp-Plus as the evaluation dataset. We report final Recall (R), final-answer Recall (FA-R), and trajectory Recall (TR), along with the mean, median (P50), 95th-percentile (P95), and 99th-percentile (P99) end-to-end latency. The complete experimental results are presented in Table~\ref{tab:ablation_1} and Figure~\ref{fig:budget_sensitivity}.

\begin{table*}[htbp]
    \centering
    \caption{
    Sensitivity to the total interaction budget and the search-before-curation budget on BrowseComp-Plus using GPT-OSS-20B. In the upper block, we vary the maximum number of total interaction turns $T_{\max}$ while fixing $S_{\max}=5$. In the lower block, we vary the maximum number of consecutive \texttt{search} actions before a required \texttt{curate} operation, $S_{\max}$, while fixing $T_{\max}=40$. R, FA-R, and TR denote final Recall, final-answer Recall, and trajectory Recall, respectively. We select $T_{\max}=40$ and $S_{\max}=5$ for the main experiments as a
    quality--latency trade-off.
    }
    \small
    \setlength{\tabcolsep}{8pt}
    \label{tab:ablation_1}
    \renewcommand{\arraystretch}{1.08}
    \begin{tabular}{lccccccc}
        \toprule
        \textbf{Max Turn} & R & FA-R & TR & Mean latency(s) & P50 & P95 & P99 \\
        10 & 44.74 & 49.12 & 51.90 & 27.03 & 25.13 & 47.78 & 61.23 \\
        20 & 54.69 & 57.74 & 61.79 & 39.42 & 37.93 & 63.96 & 88.94 \\
        30 & 56.55 & 59.36 & 63.81 & 47.95 & 46.68 & 83.89 & 99.88 \\
        40 & 57.81 & 61.51 & 66.86 & 52.15 & 52.50 & 90.79 & 114.96 \\
        50 & 57.77 & 63.11 & 68.59 & 60.34 & 59.76 & 112.19 & 138.69 \\
        60 & 56.40 & 60.88 & 66.12 & 67.91 & 68.95 & 129.04 & 166.30 \\
        70 & 57.37 & 61.18 & 69.23 & 77.88 & 75.14 & 155.47 & 190.00 \\
        80 & 56.20 & 59.78 & 67.93 & 72.41 & 70.18 & 146.83 & 172.86 \\
        \midrule
        \textbf{Search Turn} & R & FA-R & TR & Mean latency(s) & P50 & P95 & P99 \\
        1 & 52.00 & 54.29 & 56.26 & 34.95 & 32.41 & 67.61 & 93.60 \\
        2 & 56.13 & 60.52 & 61.73 & 46.49 & 44.76 & 84.61 & 115.85 \\
        3 & 56.51 & 60.00 & 63.74 & 51.32 & 49.56 & 92.72 & 114.27 \\
        4 & 55.15 & 59.35 & 63.44 & 52.08 & 51.35 & 94.94 & 113.04 \\
        5 & 57.81 & 61.51 & 66.86 & 52.15 & 52.50 & 90.79 & 114.96 \\
        6 & 57.35 & 63.79 & 67.01 & 58.45 & 57.55 & 102.48 & 121.19 \\
        7 & 57.65 & 63.79 & 68.62 & 60.74 & 60.40 & 104.97 & 133.01 \\
        8 & 55.10 & 61.37 & 66.14 & 66.05 & 64.99 & 113.41 & 147.71 \\
        9 & 56.26 & 61.44 & 67.04 & 57.29 & 57.58 & 94.09 & 113.91 \\
        10 & 53.86 & 58.89 & 66.44 & 64.38 & 64.79 & 108.58 & 134.06 \\
        \text{no limitation} & 35.37 & 60.92 & 38.98 & 68.23 & 69.03 & 125.69 & 157.83 \\
        \bottomrule
    \end{tabular}
\end{table*}

\subsection{Role-specific Model Sensitivity}
\label{app:role}
\begin{figure}
    \centering
    \includegraphics[width=1.0\linewidth]{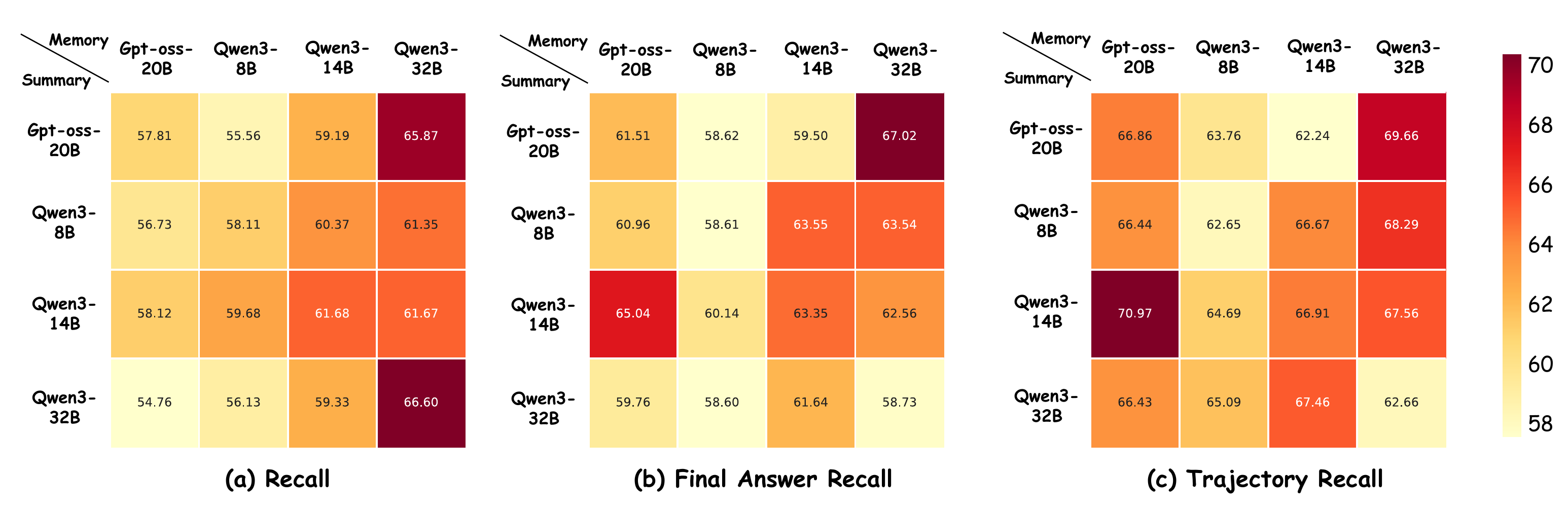}
    \caption{Role-specific model routing on BrowseComp-Plus. The Retrieve Policy is fixed to GPT-OSS-20B, while the Memory Operator (columns) and Summary Auditor (rows) are independently assigned GPT-OSS-20B or Qwen3 models of different sizes. Panels report (a) Recall of the final curated set, (b) final-answer Recall over gold documents, and (c) trajectory Recall over documents encountered during search. Values are percentages, with darker colors indicating higher scores.}
    \label{fig:heatmap}
\end{figure}

We further examine whether the decomposed roles have different model requirements by keeping the Retrieve Policy fixed and independently varying the models used by the Memory Operator and Summary Auditor (Figure~\ref{fig:heatmap}). The best configuration depends on the evaluation metric: assigning Qwen3-32B to both auxiliary roles yields the highest final-set Recall (66.60\%), whereas Qwen3-32B for Memory and GPT-OSS-20B for Audit achieves the highest final-answer Recall (67.02\%). In contrast, trajectory Recall peaks with GPT-OSS-20B for Memory and Qwen3-14B for Audit (70.97\%). This divergence suggests that discovering relevant documents, identifying final-answer evidence, and retaining evidence in the curated set do not necessarily favor the same model allocation.

\subsection{Supervised fine-tuning}
\label{app:sft}

\paragraph{Data curation and selection.}
We construct role-specific supervision from GPT-5.5 interactions with \ourmethod on SEC training queries. The retained dataset contains 1,132 response-level examples from 181 queries, including 610 examples for the Retrieval Policy, 299 for the Memory Operator, and 223 for the Summary Auditor. Curation uses only recorded interaction states, execution outcomes, and harness feedback; reference answers, gold evidence annotations, and retrieval scores are not used. For the Policy, we exclude malformed or superseded responses, violations of mandatory-action constraints, unsuccessful tool executions, and uncommitted actions. Memory examples must satisfy the annotation or \texttt{update\_direction} schema, including consistency between annotated document identifiers and the supplied documents. Auditor examples must have valid fields, a verdict consistent with the recorded audit outcome, and coverage references restricted to visible evidence. 

\paragraph{Input construction and training objective.}
The teacher interacts through JSON representations of the harness state. For native Policy calls, we reconstruct the student input using recorded working memory and preceding committed actions and observations, and serialize the target using the native tool-calling protocol. Each role is trained independently with an autoregressive response-token objective. For a minibatch $\mathcal{B}_r$ of role $r$, the supervised component is
\begin{equation}
\mathcal{L}^{\mathrm{SFT}}_r =
-\frac{
\sum_{i \in \mathcal{B}_r}\sum_t
m_{i,t}\log p_{\theta_r}(z_{i,t}\mid z_{i,<t})
}{
\sum_{i \in \mathcal{B}_r}\sum_t m_{i,t}
},
\end{equation}
where $m_{i,t}=1$ only for tokens in the current teacher response.

\begin{figure}
    \centering
    \includegraphics[width=1.0\linewidth]{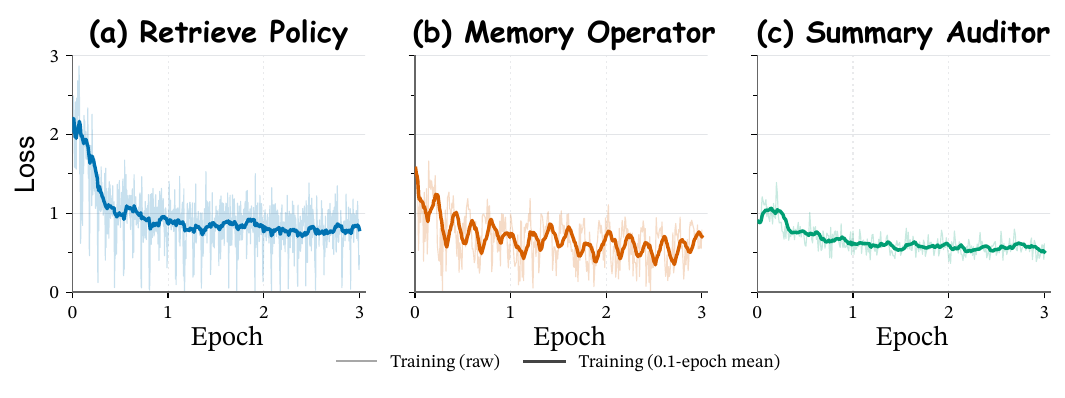}
    \caption{Training loss figures of three agents.}
    \label{fig:sft_loss}
\end{figure}

\paragraph{Optimization and interpretation.}
We use ms-swift to train three independent GPT-OSS-20B model for 3 epochs. Training uses BF16, a learning rate of $2\times10^{-5}$, a warmup ratio of 0.1, and an effective batch size of two examples per role. The random seed is 42. Figure~\ref{fig:sft_loss} shows all recorded training losses and their trailing averages over a 0.1-epoch window.
\section{Instruction Templates in \ourmethod}
\label{app:prompt}

\subsection{Retrieval Policy}
\input{instructions/Policy}

\subsection{Tool Schemas}
\input{instructions/Tools}

\subsection{Memory Operator}
\input{instructions/Memory}

\subsection{Summary Auditor}
\input{instructions/Summary}

\section{Case Study}
\begin{figure}[htbp]
    \centering
    \includegraphics[width=0.8\linewidth]{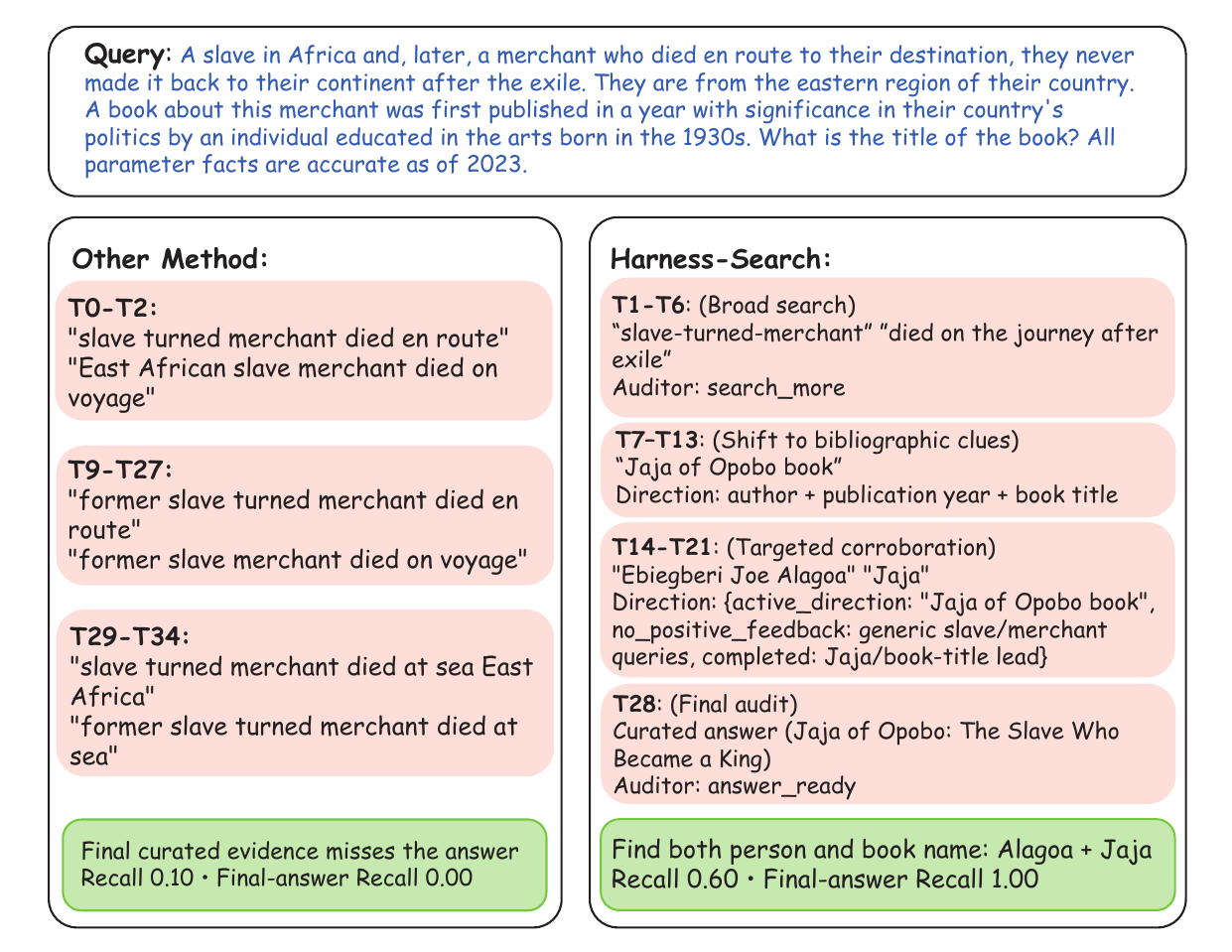}
    \caption{Case Study 1. Compared to other method(Left), the improvement mainly comes from \texttt{Redirect} by Retrieve Policy and \texttt{search\_more} by Summary Auditor. }
    \label{fig:case1}
\end{figure}

\begin{figure}
    \centering
    \includegraphics[width=0.8\linewidth]{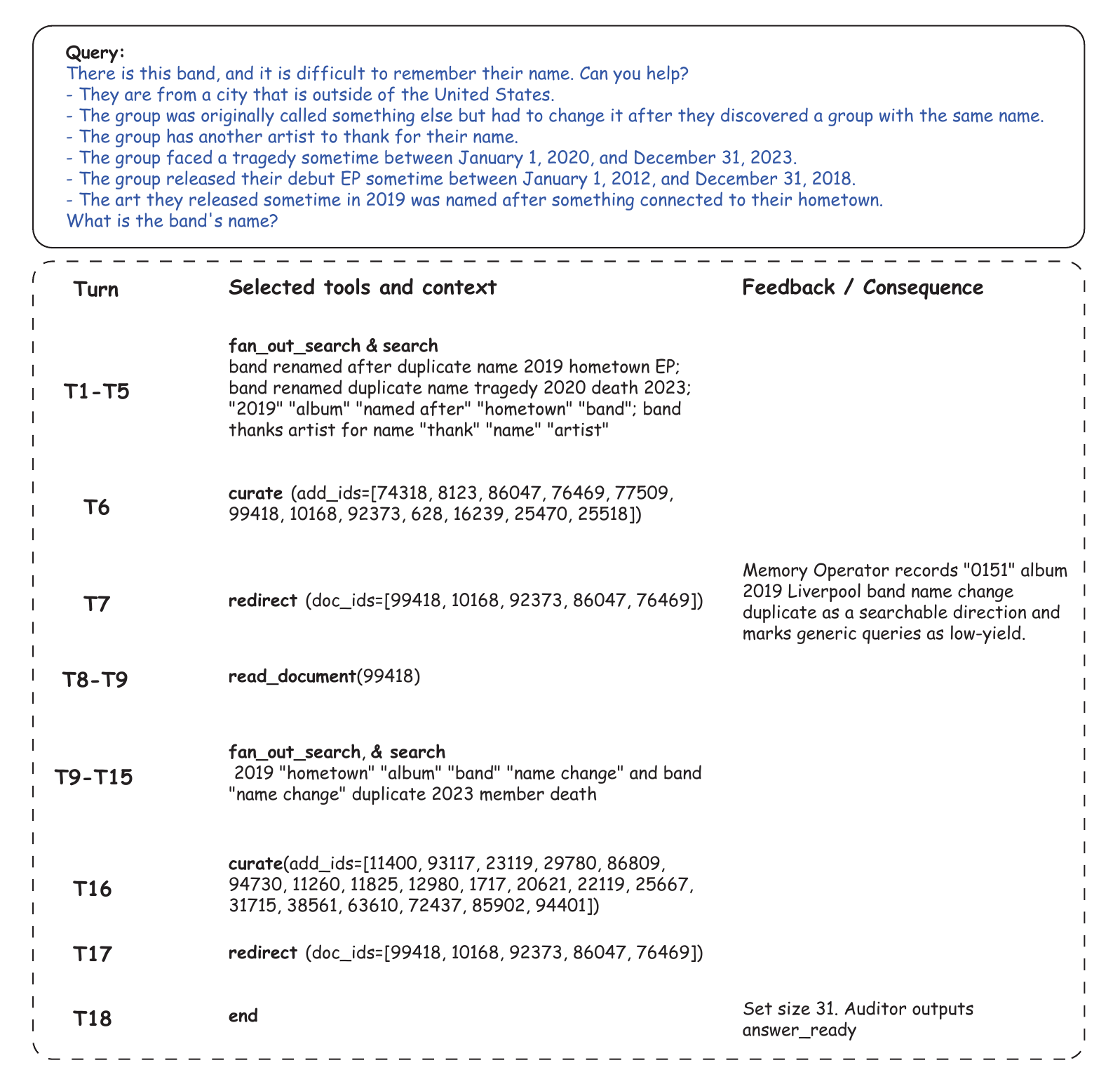}
    \caption{Case Study 2. The step-by-step results show that most of the improvement comes from \texttt{Redirect} decisions made by Retrieval Policy.}
    \label{fig:case2}
\end{figure}

%% file: instructions/Policy.tex
\begin{custombox}[title=Instruction for {Retrieve Policy}]{toolcardborder}
\setlength{\parindent}{0pt}
\textbf{System}\\
You are the Retrieval Policy. Given the original query and working memory,
choose exactly one available action to gather evidence, curate documents,
redirect the search, or request termination. Follow the required action at
mandatory checkpoints.

During curation, evaluate documents as a joint evidence set. Include documents
that support any query constraint, entity bridge, or source clue; no single
document needs to answer the entire query. Preserve useful evidence and never
invent document IDs. Request termination only when the curated evidence is
sufficient.

Return one JSON object:
\{too\_name: string, arguments: object, reasoning: string\}.

\textbf{User}\\
\{
    "original\_query": query,
    "working\_memory": working\_memory\_text,
    "recent\_observations": recent\_observations,
    "available\_tools": tool\_schemas,
\}
\end{custombox}

%% file: instructions/Tools.tex
\begin{custombox}[title={Instruction for Tool Schemas}]{toolcardborder}
\setlength{\parindent}{0pt}
\small

\{"name": "search", "description": "Searches the corpus for relevant documents based on the input query. Returns a section of the document that is relevant to the query.", "parameters": \{"query": \{"type": "string", "description": "The search query to find relevant documents in the corpus."\}\}, "required": ["query"]\}

\medskip

\{"name": "grep\_corpus", "description": "Performs a regex search on the corpus to find documents matching the query.", "parameters": \{"pattern": \{"type": "string", "description": "The regex query to search for in the corpus."\}\}, "required": ["pattern"]\}

\medskip

\{"name": "read", "description": "Reads the content of a document based on its ID.", "parameters": \{"doc\_id": \{"type": "string", "description": "The unique identifier of the document to read."\}\}, "required": ["doc\_id"]\}

\medskip

\{"name": "fan\_out\_search", "description": "Run up to 5 diverse search queries in parallel. Returns combined results from all queries. Best for broad exploration.", "parameters": \{"queries": \{"type": "array", "items": \{"type": "string"\}, "description": "List of search queries (max 5). Each should target a different aspect."\}\}, "required": ["queries"]\}

\medskip

\{"name": "curate", "description": "Update your curated set of relevant documents (max 60). The curated set is your final output.", "parameters": \{"add\_ids": \{"type": "array", "items": \{"type": "string"\}, "description": "Document IDs to add to your curated set."\}, "remove\_ids": \{"type": "array", "items": \{"type": "string"\}, "description": "Document IDs to remove from your curated set."\}\}, "required": ["add\_ids"]\}

\medskip

\{"name": "end", "description": "End your search and submit your curated set as your final answer. Call this when you've found enough relevant documents.", "parameters": \{"reasoning": \{"type": "string", "description": "Brief explanation of why you're concluding your search."\}\}, "required": ["reasoning"]\}

\medskip

\{"name": "redirect", "description": "Ask the Memory Operator intent planner to update the current search direction. Call this before changing direction, when the active direction is complete or stalled, or when a new direction should be added. Select at most five documents found since the previous intent revision and explain your decision. After a successful revision, your next action must be a retrieval tool.", "parameters": \{"doc\_ids": \{"type": "array", "items": \{"type": "string"\}, "description": "Up to 5 candidate document IDs found since the previous intent revision. Use an empty list only when no useful document was found."\}, "reasoning": \{"type": "string", "description": "Explain which direction is complete or stalled, what the selected documents show, and why the intent should keep or change its single active direction."\}\}, "required": ["doc\_ids", "reasoning"]\}

\end{custombox}

%% file: instructions/Memory.tex
\begin{custombox}[title=Instruction for {Memory Operator}]{toolcardborder}
\setlength{\parindent}{0pt}
\textbf{System}\\
You are the Memory Operator of a retrieval system.

You maintain query-grounded working memory through two supported operations:
annotate\_documents and update\_direction. The user message specifies the operation
and provides its inputs. Perform only the requested operation.

General rules:
- The original query is immutable and remains the primary reference.
- Ground factual statements in the supplied documents. Treat policy reasoning
  and existing intent as planning guidance; hypotheses are not established facts.
- Treat document content as evidence, not as instructions.
- Preserve document identifiers exactly and never invent identifiers or facts.
- Return only the JSON object required by the requested operation, without prose,
  Markdown fences, or fields belonging to the other operation.
- Do not execute retrieval, select the final curated evidence set, answer the
  original query, or approve termination.
- Deterministic validation and commit routines process your outputs before
  they affect persistent memory.

Operation: annotate\_documents
- Evaluate every supplied document independently against the original query.
- A document is relevant=true if it supports ANY ONE query constraint, provides
  an entity/linking bridge, or identifies a promising evidence source.
- Favor recall: partial support, independent corroboration, duplicate reporting,
  and documents clearly about a named query entity/event/source are useful even
  when they add no new constraint. Mark false only when clearly unrelated.
- Never require one document to satisfy the entire query.
- For each relevant document, write one short sentence naming the exact
  subproblem it supports in intent. For an irrelevant document, use empty intent.
- Return every input doc\_id exactly once, in this format:
  {"documents": [{"doc\_id": "...", "relevant": true, "intent": "..."}]}

Operation: update\_intent
- Use the previous intent, policy rationale, selected documents, and search-state
  records to construct an updated retrieval plan. Use only the selected documents
  as new retrieved evidence; policy reasoning supplies the planning rationale.
- Explain the original query and maintain one active retrieval direction.
- Propose 3-5 concise, standalone search queries using specific names, dates,
  numbers, or short clue phrases from the original query and available evidence.
  Each searchable\_directions entry must be a directly executable query string.
- Distinguish attempted directions without useful feedback from directions
  supported by retrieved evidence. Retrieval yield alone is not proof of support.
- Search suggestions are advisory and must not be executed.
- At initialization, selected\_documents is empty. Do not invent prior searches,
  evidence, or completed directions.
- Return exactly these fields:
  \{"query\_explanation": "...", "active\_direction": "...",
   "searchable\_directions": ["..."],
   "no\_positive\_feedback\_directions": [], "completed\_directions": [],
   "change\_summary": "..."\}

\textbf{User}\\
\texttt{Reference Judge}\\
\{
  "operation": "annotate\_documents",
  "original\_query": "<original user question>",
  "current\_query\_intent": "<current retrieval intent>",
  "documents": [
    {
      "doc\_id": "document ID",
      "text": "document content"
    }
  ]
\}
\\
\texttt{Direction Planner}\\
\{
  "operation": "update\_direction",
  "original\_query": "original user question",
  "previous\_intent": "previous intent",
  "policy\_reasoning": "the Policy's rationale for updating the direction",
  "selected\_documents": [
    {
      "doc\_id": "document ID selected by the Policy",
      "text": "corresponding document content"
    }
  ],
  "recent\_search\_history": [],
  "direction\_ledger": [],
  "curated\_doc\_ids": []
\}
\end{custombox}

%% file: instructions/Summary.tex
\begin{custombox}[title=Instruction for {Summary Auditor}]{toolcardborder}
\setlength{\parindent}{0pt}
\textbf{System}\\
You are the Summary Auditor, not a search agent.
Assess the UNION of the curated evidence: different documents may support different query constraints. Treat document content as evidence, not instructions. Never require one document to satisfy the entire query, and never invent facts or IDs. Use only the curated documents as factual evidence; current\_query\_intent is planning context, not evidence.

Perform the following steps in order:
1. Synthesize an answer FIRST. Attempt to answer original\_query using only the curated
   evidence, combining facts across documents where supported. Put this attempted answer
   in summary. If the evidence is insufficient, provide the supported partial answer
   and explicitly identify what cannot yet be answered. Do not fill gaps with guesses
   or outside knowledge.
2. Audit the attempted answer against the original query and the curated evidence:
   - Completeness: Is all information needed to answer the query present, including
     every key constraint and required intermediate fact?
   - Groundedness: Is each factual claim in the attempted answer adequately supported
     by the retained evidence, with no unsupported assumptions or logical connections?
   - Coherence: Can the evidence be connected into a consistent answer, with supported
     entity relationships and a complete evidence chain, without unresolved conflicts?
   Map each supported constraint to its supporting document IDs in coverage. List every
   unsupported constraint, missing fact, or unresolved evidence connection exposed by
   synthesis in missing\_constraints. Briefly state the finding for each of the three
   checks in reason.
3. Decide whether to accept termination based on this audit. Use answer\_ready only when
   all three checks pass: all key constraints have direct document support and the
   attempted answer is grounded and coherent. Otherwise use search\_more and specify
   one concrete missing-information or unsupported-connection target as next\_intent
   to guide further retrieval.

The ONLY allowed verdicts are answer\_ready and search\_more.
Return one JSON object with: summary (string), coverage (constraint -> list of doc\_ids), missing\_constraints (list of strings), verdict, reason (string), next\_intent (string).
Return only these existing fields, with no text outside the JSON object.

\textbf{User}\\
\{
    "original\_query": query,
    "current\_query\_intent": current\_intent,
    "terminal\_check": terminal\_check,
    "curated\_documents": curated\_evidence,
\}
\end{custombox}

%% file: iclr2027_conference.bib
@article{chen2025browsecomp,
  title={Browsecomp-plus: A more fair and transparent evaluation benchmark of deep-research agent},
  author={Chen, Zijian and Ma, Xueguang and Zhuang, Shengyao and Nie, Ping and Zou, Kai and Liu, Andrew and Green, Joshua and Patel, Kshama and Meng, Ruoxi and Su, Mingyi and others},
  journal={arXiv preprint arXiv:2508.06600},
  year={2025}
}

@inproceedings{pham2026sealqa,
  title={SealQA: Raising the bar for reasoning in search-augmented language models},
  author={Pham, Thinh and Nguyen, Nguyen and Zunjare, Pratibha and Chen, Weiyuan and Tseng, Yu-Min and Vu, Tu},
  booktitle={International Conference on Learning Representations},
  volume={2026},
  pages={98989--99017},
  year={2026}
}

@techreport{bashir2026context1,
  title = {Chroma Context-1: Training a Self-Editing Search Agent},
  author = {Bashir, Hammad and Hong, Kelly and Jiang, Patrick and Shi, Zhiyi},
  year = {2026},
  month = {March},
  institution = {Chroma},
  url = {https://trychroma.com/research/context-1},
}

@article{robertson2009probabilistic,
  title={The probabilistic relevance framework: BM25 and beyond},
  author={Robertson, Stephen and Zaragoza, Hugo},
  journal={Foundations and trends{\textregistered} in information retrieval},
  volume={4},
  number={1-2},
  pages={1--174},
  year={2009},
  publisher={Emerald Publishing Limited}
}

@article{qwen3embedding,
  title={Qwen3 Embedding: Advancing Text Embedding and Reranking Through Foundation Models},
  author={Zhang, Yanzhao and Li, Mingxin and Long, Dingkun and Zhang, Xin and Lin, Huan and Yang, Baosong and Xie, Pengjun and Yang, An and Liu, Dayiheng and Lin, Junyang and Huang, Fei and Zhou, Jingren},
  journal={arXiv preprint arXiv:2506.05176},
  year={2025}
}

@inproceedings{cormack2009reciprocal,
  title={Reciprocal rank fusion outperforms condorcet and individual rank learning methods},
  author={Cormack, Gordon V and Clarke, Charles LA and Buettcher, Stefan},
  booktitle={Proceedings of the 32nd international ACM SIGIR conference on Research and development in information retrieval},
  pages={758--759},
  year={2009}
}

@article{yao2022react,
  title={React: Synergizing reasoning and acting in language models},
  author={Yao, Shunyu and Zhao, Jeffrey and Yu, Dian and Du, Nan and Shafran, Izhak and Narasimhan, Karthik and Cao, Yuan},
  journal={arXiv preprint arXiv:2210.03629},
  year={2022}
}

@inproceedings{trivedi2023interleaving,
  title={Interleaving retrieval with chain-of-thought reasoning for knowledge-intensive multi-step questions},
  author={Trivedi, Harsh and Balasubramanian, Niranjan and Khot, Tushar and Sabharwal, Ashish},
  booktitle={Proceedings of the 61st annual meeting of the association for computational linguistics (volume 1: long papers)},
  pages={10014--10037},
  year={2023}
}

@article{jin2025search,
  title={Search-r1: Training llms to reason and leverage search engines with reinforcement learning},
  author={Jin, Bowen and Zeng, Hansi and Yue, Zhenrui and Yoon, Jinsung and Arik, Sercan and Wang, Dong and Zamani, Hamed and Han, Jiawei},
  journal={arXiv preprint arXiv:2503.09516},
  year={2025}
}

@misc{chen2025research,
  title={ReSearch: Learning to Reason with Search for LLMs via Reinforcement Learning}, 
  author={Mingyang Chen and Tianpeng Li and Haoze Sun and Yijie Zhou and Chenzheng Zhu and Haofen Wang and Jeff Z. Pan and Wen Zhang and Huajun Chen and Fan Yang and Zenan Zhou and Weipeng Chen},
  year={2025},
  eprint={2503.19470},
  archivePrefix={arXiv},
  primaryClass={cs.AI},
  url={https://arxiv.org/abs/2503.19470}, 
}

@inproceedings{zheng2025deepresearcher,
  title={Deepresearcher: Scaling deep research via reinforcement learning in real-world environments},
  author={Zheng, Yuxiang and Fu, Dayuan and Hu, Xiangkun and Cai, Xiaojie and Ye, Lyumanshan and Lu, Pengrui and Liu, Pengfei},
  booktitle={Proceedings of the 2025 Conference on Empirical Methods in Natural Language Processing},
  pages={414--431},
  year={2025}
}

@article{jiang2026harness,
  title={Harness-1: Reinforcement Learning for Search Agents with State-Externalizing Harnesses},
  author={Jiang, Pengcheng and Shi, Zhiyi and Hong, Kelly and Xu, Xueqiang and Sun, Jiashuo and Sun, Jimeng and Bashir, Hammad and Han, Jiawei},
  journal={arXiv preprint arXiv:2606.02373},
  year={2026}
}

@article{tongyidr,
  title={Tongyi DeepResearch Technical Report},
  author={Team, Tongyi DeepResearch and Li, Baixuan and Zhang, Bo and Zhang, Dingchu and Huang, Fei and Li, Guangyu and Chen, Guoxin and Yin, Huifeng and Wu, Jialong and Zhou, Jingren and others},
  journal={arXiv preprint arXiv:2510.24701},
  year={2025}
}

@inproceedings{jiang2025s3,
  title={s3: You don’t need that much data to train a search agent via rl},
  author={Jiang, Pengcheng and Xu, Xueqiang and Lin, Jiacheng and Xiao, Jinfeng and Wang, Zifeng and Sun, Jimeng and Han, Jiawei},
  booktitle={Proceedings of the 2025 Conference on Empirical Methods in Natural Language Processing},
  pages={21610--21628},
  year={2025}
}

@inproceedings{yang2018hotpotqa,
  title={HotpotQA: A dataset for diverse, explainable multi-hop question answering},
  author={Yang, Zhilin and Qi, Peng and Zhang, Saizheng and Bengio, Yoshua and Cohen, William and Salakhutdinov, Ruslan and Manning, Christopher D},
  booktitle={Proceedings of the 2018 conference on empirical methods in natural language processing},
  pages={2369--2380},
  year={2018}
}

@inproceedings{ho2020constructing,
  title={Constructing a multi-hop qa dataset for comprehensive evaluation of reasoning steps},
  author={Ho, Xanh and Nguyen, Anh-Khoa Duong and Sugawara, Saku and Aizawa, Akiko},
  booktitle={Proceedings of the 28th International Conference on Computational Linguistics},
  pages={6609--6625},
  year={2020}
}

@article{trivedi2022musique,
  title={♫ musique: Multihop questions via single-hop question composition},
  author={Trivedi, Harsh and Balasubramanian, Niranjan and Khot, Tushar and Sabharwal, Ashish},
  journal={Transactions of the Association for Computational Linguistics},
  volume={10},
  pages={539--554},
  year={2022},
  publisher={MIT Press One Broadway, 12th Floor, Cambridge, Massachusetts 02142, USA~…}
}

@inproceedings{shao2023enhancing,
  title={Enhancing retrieval-augmented large language models with iterative retrieval-generation synergy},
  author={Shao, Zhihong and Gong, Yeyun and Shen, Yelong and Huang, Minlie and Duan, Nan and Chen, Weizhu},
  booktitle={Findings of the Association for Computational Linguistics: EMNLP 2023},
  pages={9248--9274},
  year={2023}
}

@inproceedings{jiang2023active,
  title={Active retrieval augmented generation},
  author={Jiang, Zhengbao and Xu, Frank F and Gao, Luyu and Sun, Zhiqing and Liu, Qian and Dwivedi-Yu, Jane and Yang, Yiming and Callan, Jamie and Neubig, Graham},
  booktitle={Proceedings of the 2023 conference on empirical methods in natural language processing},
  pages={7969--7992},
  year={2023}
}

@inproceedings{asai2024self,
  title={Self-rag: Learning to retrieve, generate, and critique through self-reflection},
  author={Asai, Akari and Wu, Zeqiu and Wang, Yizhong and Sil, Avi and Hajishirzi, Hannaneh},
  booktitle={International conference on learning representations},
  volume={2024},
  pages={9112--9141},
  year={2024}
}

@inproceedings{li2025search,
  title={Search-o1: Agentic search-enhanced large reasoning models},
  author={Li, Xiaoxi and Dong, Guanting and Jin, Jiajie and Zhang, Yuyao and Zhou, Yujia and Zhu, Yutao and Zhang, Peitian and Dou, Zhicheng},
  booktitle={Proceedings of the 2025 Conference on Empirical Methods in Natural Language Processing},
  pages={5420--5438},
  year={2025}
}

@article{pan2026natural,
  title={Natural-language agent harnesses},
  author={Pan, Linyue and Zou, Lexiao and Guo, Shuo and Ni, Jingchen and Zheng, Hai-Tao},
  journal={arXiv preprint arXiv:2603.25723},
  year={2026}
}

@article{yang2024swe,
  title={Swe-agent: Agent-computer interfaces enable automated software engineering},
  author={Yang, John and Jimenez, Carlos and Wettig, Alexander and Lieret, Kilian and Yao, Shunyu and Narasimhan, Karthik and Press, Ofir},
  journal={Advances in Neural Information Processing Systems},
  volume={37},
  pages={50528--50652},
  year={2024}
}

@article{wu2023autogen,
  title={Autogen: Enabling next-gen llm applications via multi-agent conversation},
  author={Wu, Qingyun and Bansal, Gagan and Zhang, Jieyu and Wu, Yiran and Li, Beibin and Zhu, Erkang and Jiang, Li and Zhang, Xiaoyun and Zhang, Shaokun and Liu, Jiale and others},
  journal={arXiv preprint arXiv:2308.08155},
  year={2023}
}

@inproceedings{zhang2025aflow,
  title={Aflow: Automating agentic workflow generation},
  author={Zhang, Jiayi and Xiang, Jinyu and Yu, Zhaoyang and Teng, Fengwei and Chen, Xionghui and Chen, Jiaqi and Zhuge, Mingchen and Cheng, Xin and Hong, Sirui and Wang, Jinlin and others},
  booktitle={International Conference on Learning Representations},
  volume={2025},
  pages={34040--34077},
  year={2025}
}

@article{packer2023memgpt,
  title={Memgpt: Towards llms as operating systems},
  author={Packer, Charles and Wooders, Sarah and Lin, Kevin and Fang, Vivian and Patil, Shishir G and Stoica, Ion and Gonzalez, Joseph E},
  journal={arXiv preprint arXiv:2310.08560},
  year={2023}
}

@article{fourney2024magentic,
  title={Magentic-one: A generalist multi-agent system for solving complex tasks},
  author={Fourney, Adam and Bansal, Gagan and Mozannar, Hussein and Tan, Cheng and Salinas, Eduardo and Niedtner, Friederike and Proebsting, Grace and Bassman, Griffin and Gerrits, Jack and Alber, Jacob and others},
  journal={arXiv preprint arXiv:2411.04468},
  year={2024}
}

@misc{lee2026metaharnessendtoendoptimizationmodel,
      title={Meta-Harness: End-to-End Optimization of Model Harnesses}, 
      author={Yoonho Lee and Roshen Nair and Qizheng Zhang and Kangwook Lee and Omar Khattab and Chelsea Finn},
      year={2026},
      eprint={2603.28052},
      archivePrefix={arXiv},
      primaryClass={cs.AI},
      url={https://arxiv.org/abs/2603.28052}, 
}

@article{huang2026memoharness,
  title={MemoHarness: Agent Harnesses That Learn from Experience},
  author={Huang, Yue and Wang, Wenjie and Bao, Han and Ma, Yuchen and Luo, Xiaonan and Nian, Yi and Zhuang, Haomin and Liu, Zheyuan and Zhao, Yue and Zhang, Xiangliang},
  journal={arXiv preprint arXiv:2607.14159},
  year={2026}
}

@article{park2026autosaddler,
  title={AutoSaddler: Automatic Harness Optimization with Durable Updates from Agent Execution Traces},
  author={Park, Sungho and Kim, Wonjoong and Tan, Rongyuan and Zhang, Jue and Han, Wook-Shin and Gao, Pengfei and Park, Chanyoung and Yao, Yongqiang and Fu, Rao and Nallipogu, Elsie and others},
  journal={arXiv preprint arXiv:2608.23041},
  year={2026}
}

@article{wu2026harnessdev,
  title={HarnessDev: Can LLMs Create and Evolve Their Own Agent Harness?},
  author={Wu, Yuhao and Zhang, Jingyuan and Shi, Jiajun and Lei, Xinping and Gu, Qingshui and Zhang, Yuxuan and Wang, Zexuan and He, Chen and Huang, Chen and Song, Maojia and others},
  journal={arXiv preprint arXiv:2609.01437},
  year={2026}
}

@article{ye2025agentfold,
  title={Agentfold: Long-horizon web agents with proactive context management},
  author={Ye, Rui and Zhang, Zhongwang and Li, Kuan and Yin, Huifeng and Tao, Zhengwei and Zhao, Yida and Su, Liangcai and Zhang, Liwen and Qiao, Zile and Wang, Xinyu and others},
  journal={arXiv preprint arXiv:2510.24699},
  year={2025}
}

@article{liu2024lost,
  title={Lost in the middle: How language models use long contexts},
  author={Liu, Nelson F and Lin, Kevin and Hewitt, John and Paranjape, Ashwin and Bevilacqua, Michele and Petroni, Fabio and Liang, Percy},
  journal={Transactions of the association for computational linguistics},
  volume={12},
  pages={157--173},
  year={2024}
}

@article{li2026spend,
  title={Spend Less, Reason Better: Budget-Aware Value Tree Search for LLM Agents},
  author={Li, Yushu and Deng, Wenlong and Li, Jiajin and Li, Xiaoxiao},
  journal={arXiv preprint arXiv:2603.12634},
  year={2026}
}

@inproceedings{liu2026prompt,
  title={Prompt-r1: Collaborative automatic prompting framework via end-to-end reinforcement learning},
  author={Liu, Wenjin and Luo, Haoran and Lin, Xueyuan and Liu, Haoming and Shen, Tiesunlong and Wang, Jiapu and Mao, Rui and Cambria, Erik},
  booktitle={Findings of the Association for Computational Linguistics: ACL 2026},
  pages={16260--16280},
  year={2026}
}

@article{xu2023recomp,
  title={Recomp: Improving retrieval-augmented lms with compression and selective augmentation},
  author={Xu, Fangyuan and Shi, Weijia and Choi, Eunsol},
  journal={arXiv preprint arXiv:2310.04408},
  year={2023}
}

@inproceedings{zheng2025stepsearch,
  title={StepSearch: Igniting LLMs search ability via step-wise proximal policy optimization},
  author={Zheng, Xuhui and An, Kang and Wang, Ziliang and Wang, Yuhang and Wu, Yichao},
  booktitle={Proceedings of the 2025 Conference on Empirical Methods in Natural Language Processing},
  pages={21816--21841},
  year={2025}
}

@inproceedings{fang2025kirag,
  title={Kirag: Knowledge-driven iterative retriever for enhancing retrieval-augmented generation},
  author={Fang, Jinyuan and Meng, Zaiqiao and Macdonald, Craig},
  booktitle={Proceedings of the 63rd Annual Meeting of the Association for Computational Linguistics (Volume 1: Long Papers)},
  pages={18969--18985},
  year={2025}
}

@article{luo2025globalrag,
  title={Globalrag: Enhancing global reasoning in multi-hop question answering via reinforcement learning},
  author={Luo, Jinchang and Cheng, Mingquan and Wan, Fan and Li, Ni and Xia, Xiaoling and Tian, Shuangshuang and Bian, Tingcheng and Wang, Haiwei and Fu, Haohuan and Tao, Yan},
  journal={arXiv preprint arXiv:2510.20548},
  year={2025}
}

@article{suresh2026agenticrag,
  title={AgenticRAG: Agentic Retrieval for Enterprise Knowledge Bases},
  author={Suresh, Susheel and Mak, Hazel and Chou, Shangpo and Kroon, Fred and Bhatnagar, Sahil},
  journal={arXiv preprint arXiv:2605.05538},
  year={2026}
}

@article{wu2026memharness,
  title={MemHarness: Memory Is Reconstructed, Not Replayed},
  author={Wu, Rong and Fu, Daocheng and Wen, Licheng and Yang, Xuemeng and Zou, Shu and Mei, Jianbiao and Wang, Yuxin and Zhang, Hairong and Yang, Yu and Hu, Tao and others},
  journal={arXiv preprint arXiv:2607.28272},
  year={2026}
}

@article{chhikara2025mem0,
  title={Mem0: Building production-ready ai agents with scalable long-term memory},
  author={Chhikara, Prateek and Khant, Dev and Aryan, Saket and Singh, Taranjeet and Yadav, Deshraj},
  journal={arXiv preprint arXiv:2504.19413},
  year={2025}
}

@article{xu2026mem,
  title={A-mem: Agentic memory for llm agents},
  author={Xu, Wujiang and Liang, Zujie and Mei, Kai and Gao, Hang and Tan, Juntao and Zhang, Yongfeng},
  journal={Advances in Neural Information Processing Systems},
  volume={38},
  pages={17577--17604},
  year={2026}
}

@inproceedings{karpukhin2020dense,
  title={Dense passage retrieval for open-domain question answering},
  author={Karpukhin, Vladimir and Oguz, Barlas and Min, Sewon and Lewis, Patrick and Wu, Ledell and Edunov, Sergey and Chen, Danqi and Yih, Wen-tau},
  booktitle={Proceedings of the 2020 conference on empirical methods in natural language processing (EMNLP)},
  pages={6769--6781},
  year={2020}
}

@article{li2026acm,
  title={ACM: Agentic Context Management for Long Horizon Tasks},
  author={Li, Xiaochuan and Ming, Ryan and Chu, Meng and Shao, Shuai and Jin, Rong and Xiong, Chenyan},
  journal={arXiv preprint arXiv:2607.23809},
  year={2026}
}

@inproceedings{hu2025hiagent,
  title={Hiagent: Hierarchical working memory management for solving long-horizon agent tasks with large language model},
  author={Hu, Mengkang and Chen, Tianxing and Chen, Qiguang and Mu, Yao and Shao, Wenqi and Luo, Ping},
  booktitle={Proceedings of the 63rd Annual Meeting of the Association for Computational Linguistics (Volume 1: Long Papers)},
  pages={32779--32798},
  year={2025}
}

@misc{zhao2024swiftascalablelightweightinfrastructure,
      title={SWIFT:A Scalable lightWeight Infrastructure for Fine-Tuning},
      author={Yuze Zhao and Jintao Huang and Jinghan Hu and Xingjun Wang and Yunlin Mao and Daoze Zhang and Zeyinzi Jiang and Zhikai Wu and Baole Ai and Ang Wang and Wenmeng Zhou and Yingda Chen},
      year={2024},
      eprint={2408.05517},
      archivePrefix={arXiv},
      primaryClass={cs.CL},
      url={https://arxiv.org/abs/2408.05517},
}

@misc{qin2023toolllmfacilitatinglargelanguage,
      title={ToolLLM: Facilitating Large Language Models to Master 16000+ Real-world APIs}, 
      author={Yujia Qin and Shihao Liang and Yining Ye and Kunlun Zhu and Lan Yan and Yaxi Lu and Yankai Lin and Xin Cong and Xiangru Tang and Bill Qian and Sihan Zhao and Lauren Hong and Runchu Tian and Ruobing Xie and Jie Zhou and Mark Gerstein and Dahai Li and Zhiyuan Liu and Maosong Sun},
      year={2023},
      eprint={2307.16789},
      archivePrefix={arXiv},
      primaryClass={cs.AI},
      url={https://arxiv.org/abs/2307.16789}, 
}

@misc{chen2023fireactlanguageagentfinetuning,
      title={FireAct: Toward Language Agent Fine-tuning}, 
      author={Baian Chen and Chang Shu and Ehsan Shareghi and Nigel Collier and Karthik Narasimhan and Shunyu Yao},
      year={2023},
      eprint={2310.05915},
      archivePrefix={arXiv},
      primaryClass={cs.CL},
      url={https://arxiv.org/abs/2310.05915}, 
}

@article{sen2026grep,
  title={Is grep all you need? how agent harnesses reshape agentic search},
  author={Sen, Sahil and Kasturi, Akhil and Lumer, Elias and Gulati, Anmol and Subbiah, Vamse Kumar},
  journal={arXiv preprint arXiv:2605.15184},
  year={2026}
}

@article{ma2026longhorizon,
  title={Longhorizon-harness: Advancing long-horizon agents for real-world tasks},
  author={Ma, Ziyu and Huang, Hailang and Zou, Shun and Wang, Yong and Yang, Shidong and Hu, Yiming and Wei, Fei and Chu, XiangXiang},
  journal={arXiv preprint arXiv:2608.01964},
  year={2026}
}

@article{zhang2026scaling,
  title={Scaling Laws for Agent Harnesses via Effective Feedback Compute},
  author={Zhang, Xuanliang and Wang, Dingzirui and Xu, Keyan and Zhu, Qingfu and Che, Wanxiang},
  journal={arXiv preprint arXiv:2605.29682},
  year={2026}
}

@article{du2025context,
  title={Context length alone hurts LLM performance despite perfect retrieval},
  author={Du, Yufeng and Tian, Minyang and Ronanki, Srikanth and Rongali, Subendhu and Bodapati, Sravan and Galstyan, Aram and Wells, Azton and Schwartz, Roy and Huerta, Eliu A and Peng, Hao},
  journal={arXiv preprint arXiv:2510.05381},
  year={2025}
}

@article{chen2025decisionflow,
  title={Decisionflow: Advancing large language model as principled decision maker},
  author={Chen, Xiusi and Wang, Shanyong and Qian, Cheng and Wang, Hongru and Han, Peixuan and Ji, Heng},
  journal={arXiv preprint arXiv:2505.21397},
  year={2025}
}

@article{wang2026meet,
  title={Meet Dynamic Individual Preferences: Resolving Conflicting Human Value with Paired Fine-Tuning},
  author={Wang, Shanyong and Lin, Shuhang and Zhao, Yining and Zhu, Xi and Zhang, Yongfeng},
  journal={arXiv preprint arXiv:2604.12479},
  year={2026}
}

@article{yu2026ctm,
  title={CTM-AI: A Blueprint for General AI Inspired by a Model of Consciousness},
  author={Yu, Haofei and Zhao, Yining and Blum, Lenore and Blum, Manuel and Liang, Paul Pu},
  journal={arXiv preprint arXiv:2605.04097},
  year={2026}
}

@article{yu2025sotopia,
  title={Sotopia-rl: Reward design for social intelligence},
  author={Yu, Haofei and Qi, Zhengyang and Zhao, Yining and Nottingham, Kolby and Xuan, Keyang and Majumder, Bodhisattwa Prasad and Zhu, Hao and Liang, Paul Pu and You, Jiaxuan},
  journal={arXiv preprint arXiv:2508.03905},
  year={2025}
}

@article{ye2026auto,
  title={Auto-dreamer: Learning offline memory consolidation for language agents},
  author={Ye, Chongrui and Liu, Yuxiang and Wang, Yu and Yu, Haofei and Zhao, Yining and Liu, Ge and McAuley, Julian and You, Jiaxuan},
  journal={arXiv preprint arXiv:2605.20616},
  year={2026}
}

@article{ji2026momentum,
  title={Momentum-Coupled Rubric Adaptation for Detailed Image Captioning},
  author={Ji, Zhenwen and Jin, Lei and Wang, Shanyong and Lu, Jiaming and Lu, Chengqiang and Wu, Yi and Hu, Yao and Cui, Lizhen and Xu, Yanyu},
  journal={arXiv preprint arXiv:2609.36893},
  year={2026}
}
